\documentclass[preprint,12pt,authoryear]{elsarticle}

\usepackage[utf8]{inputenc}
\usepackage[T1]{fontenc}

\usepackage{amsmath,amssymb}

\usepackage{graphicx}
\graphicspath{{./}{figures/}}

\usepackage{booktabs}
\usepackage{tabularx}
\usepackage{adjustbox}
\usepackage{multirow}
\usepackage{array}

\usepackage{xcolor}
\usepackage{capt-of}
\usepackage{microtype}
\usepackage[section]{placeins}

\usepackage{algorithm}
\usepackage{algpseudocode}

\usepackage{tikz}
\usetikzlibrary{arrows.meta, positioning, shapes.geometric}

\usepackage{float}

\usepackage{colortbl}

\usepackage[colorlinks=true,linkcolor=blue,citecolor=blue,urlcolor=blue]{hyperref}

\usepackage{cleveref}

\definecolor{modeblue}{RGB}{30,100,180}
\definecolor{modegray}{RGB}{100,100,100}

\journal{Engineering Applications of Artificial Intelligence}

\begin{document}

\begin{frontmatter}

\title{HAAS: A Policy-Aware Framework for Adaptive Task Allocation Between
       Humans and Artificial Intelligence Systems\tnoteref{t1}}

\tnotetext[t1]{This work has been submitted to Engineering Applications of
Artificial Intelligence for possible open-access publication.}

%% ---- Authors --------------------------------------------------
\author[upv]{Vicente Pelechano\corref{cor1}}
\ead{pele@vrain.upv.es}

\author[upv]{Antoni Mestre}
\ead{anmesgas@upv.es}

\author[upv]{Manoli Albert}
\ead{malbert@vrain.upv.es}

\author[uv]{Miriam Gil}
\ead{miriam.gil@uv.es}

\cortext[cor1]{Corresponding author. Camino de Vera s/n, Universitat
Politècnica de València, Valencia, Spain, E-46022.}

\affiliation[upv]{organization={Valencian Research Institute for Artificial
Intelligence (VRAIN), Universitat Polit\`ecnica de Val\`encia},
                  addressline={Camino de Vera s/n},
                  city={Valencia},
                  country={Spain}}

\affiliation[uv]{organization={Departament d'Inform\`atica,
Universitat de Val\`encia},
                 city={Valencia},
                 country={Spain}}

%% ---- Abstract -------------------------------------------------
\begin{abstract}
Deciding how to distribute work between humans and artificial intelligence
systems is a central challenge in organisational design.
Most existing approaches treat this as a binary choice, yet the operational
reality is richer: humans and artificial intelligence systems routinely share
tasks or take complementary roles depending on context, fatigue, and the stakes
involved.
Governing that distribution --- balancing efficiency, oversight, and human
capability --- remains an open problem.

This paper presents Human--Artificial-Intelligence Adaptive Symbiosis
(\textsc{HAAS}), an implemented intelligent allocation framework for
adaptive task allocation in software engineering and manufacturing.
The framework combines two coupled artificial intelligence components:
a rule-based expert system that enforces governance constraints before
any learning occurs, and a contextual-bandit learner that selects among
feasible collaboration modes from outcome feedback.
Task--agent fit is represented through five auditable cognitive dimensions
and a five-mode autonomy spectrum --- from human-only to fully autonomous
--- embedded in a reproducible benchmark spanning software-engineering and
manufacturing domains.

Three empirical findings emerge.
First, governance is not a binary switch but a tunable design variable:
tighter constraints predictably convert autonomous artificial intelligence
assignments into supervised collaborations, with domain-specific costs and
benefits.
Second, in manufacturing, stronger governance can \emph{improve} operational
performance and reduce fatigue simultaneously --- a workload-buffering effect
that runs counter to the usual framing of governance as pure overhead.
Third, no single governance setting dominates across all tested contexts;
moderate governance becomes increasingly competitive as the learner
accumulates experience within the governed action space.
Together, these findings position \textsc{HAAS} as a pre-deployment workbench
for comparing and inspecting human--artificial-intelligence allocation policies
before organisational commitment.
\end{abstract}

%% ---- Keywords -------------------------------------------------

\end{frontmatter}

%% ================================================================
\section{Introduction}
\label{sec:introduction}
%% ================================================================

As AI capabilities are embedded more deeply into organisational workflows,
work design increasingly depends on decisions about how tasks, authority, and
responsibility should be distributed between human agents and AI systems.
This is a long-standing question in human--computer interaction and
operations management, but it now arises in a broader and more operationally
consequential form because AI is no longer confined to isolated tools.
Contemporary practice still often frames the issue as a binary choice ---
either a human performs a task, or an AI does --- yet the operational
reality is considerably richer.
A software engineer may lead the design of an algorithm while an AI assistant
generates candidate implementations; a manufacturing technician may supervise
a cobot that executes repetitive assembly steps while retaining veto authority
over safety-critical decisions.
These \emph{shared} execution modes are poorly captured by existing task
allocation frameworks, which rarely model governance constraints, human
fatigue, or the progressive erosion of human skill that occurs with sustained AI
delegation.

Critically, the problem of Human--AI task allocation is not simply one of
optimising throughput.
Regulatory frameworks such as the EU AI Act~\citep{EUAIAct2024} impose
human oversight requirements for high-risk applications, and collaborative
robotics standards specify explicit authority boundaries in shared
workspaces~\citep{ISO10218,ISO15066}.
Sustained AI delegation also risks automation complacency and skill
erosion~\citep{Bainbridge1983,Parasuraman2010b}, undermining the human
capacity to detect failures and recover from disruptions.
And simply adding a human review step does not guarantee better outcomes: a
meta-analysis of 106 experiments found that Human--AI combinations often
underperform the better standalone agent, especially on decision
tasks~\citep{Vaccaro2024}.
The design question \textsc{HAAS} addresses is therefore how to
\emph{govern} task allocation so that efficiency, human oversight, and
capability-retention requirements can be audited, tuned, and evaluated
jointly under explicit trade-offs.

This paper introduces \textsc{HAAS} (Human--AI Adaptive Symbiosis),
an intelligent allocation framework that implements governed
subtask allocation and packages it as a reproducible benchmarking engine for
applied Human--AI work design.
\textsc{HAAS} makes four primary contributions:

\begin{enumerate}
  \item \textbf{A task-representation instrument for applied allocation.}
    Each subtask is characterised along five auditable dimensions ---
    repetitiveness, technical depth, creativity, ambiguity, and human
    interaction --- yielding a continuous \emph{AI affinity} signal grounded
    in cognitive task analysis and suitable for operational allocation.

  \item \textbf{A five-mode execution model for expert work.}
    Rather than a binary human/AI assignment, \textsc{HAAS} defines five
    collaboration modes (\textsc{Human-Only}, \textsc{Copilot}, \textsc{Peer},
    \textsc{Supervised}, \textsc{Autonomous}) that cover the full range from
    complete human autonomy to complete AI delegation, including three
    intermediate \emph{shared} modes with explicit human and AI participation
    shares.

  \item \textbf{A policy-aware allocation engine with human-state dynamics.}
    Allocation decisions are produced by a contextual bandit that selects
    among feasible collaboration modes from outcome feedback, subject to a
    governance layer (\texttt{PolicyEngine}) that enforces organisational
    constraints such as autonomy caps, mandatory human validation steps, and
    safety-critical overrides. Human-state dynamics --- fatigue, trust, and
    deskilling --- are embedded directly in the reward signal, so the engine
    adapts to the evolving human capacity over time.

  \item \textbf{A reproducible cross-domain simulation benchmark.}
    The allocation engine is packaged as a configurable benchmarking artefact
    spanning software engineering and manufacturing domains, enabling
    comparative evaluation of learned, heuristic, and fixed policies under
    explicit governance contracts before organisational deployment.
\end{enumerate}

Taken together, these contributions make \textsc{HAAS} both a framework for
designing governance-constrained Human--AI work and a reproducible workbench
for evaluating it.
We therefore examine three empirical questions:
(\emph{i})~which allocation strategy performs best in the audited standard
scenarios? (\emph{ii})~how does governance intensity change overall
performance and the mix of collaboration modes? And (\emph{iii})~how well do
effective governance settings and learned allocation patterns transfer across
scenarios, domains, and longer horizons?

From an expert-systems perspective, the \texttt{PolicyEngine} is consistent
with classical forward-chaining knowledge bases~\citep{Turban1992,Giarratano1994,Liao2005},
augmented with contextual-bandit learning within governance-defined bounds
(details in \Cref{sec:framework:engine}).
The remainder of the paper situates \textsc{HAAS} in the literature,
develops the framework and benchmark, reports empirical results, and closes
with implications and limitations.

%% ================================================================
\section{Background and Related Work}
\label{sec:background}
%% ================================================================

Three strands of prior work are especially important for \textsc{HAAS}:
research on Human--AI teaming clarifies the allocation problem, bandit-based
learning provides the adaptive mechanism, and governance-oriented AI research
motivates the explicit rule layer that constrains learning.

\subsection{Human--AI Teaming and Task Allocation}
\label{sec:bg:teaming}

Understanding how to allocate work between humans and AI systems requires
situating \textsc{HAAS} within two converging bodies of literature: the
classical human--automation tradition, which established the conceptual
foundations for function allocation, and the empirical literature on
AI-augmented knowledge work, which quantifies the productive impact of
contemporary AI tools in real workflows.

The theoretical roots of Human--AI work allocation trace back to
\citet{Fitts1951}, whose MABA-MABA lists (``Men Are Better At / Machines Are
Better At'') established the principle of comparative advantage as the basis
for function allocation between humans and machines.
\citet{Sheridan1992} later formalised a \emph{levels of automation} scale
ranging from full human control to full computer control, and
\citet{Parasuraman2000} extended this into a four-stage taxonomy covering
information acquisition, information analysis, decision selection, and action
implementation, arguing that appropriate automation levels depend on system
reliability and human workload.
Empirical work by \citet{Endsley1999} and \citet{Kaber2004} showed that
intermediate automation levels often optimise both performance and situation
awareness, providing an early evidence base for the graduated autonomy
approach we operationalise here.

Human--AI teaming research has extended these ideas to AI-augmented
knowledge work and machines-as-teammates agendas~\citep{Cummings2014,McNeese2018,Seeber2020},
emphasising the role of trust, situation awareness, and complementarity
between human and AI capabilities~\citep{Bansal2021,Hemmer2021}.
Design guidelines for effective Human--AI interaction~\citep{Amershi2019}
have further codified principles such as supporting efficient correction
and making clear what the system can do.
At the same time, achieving genuine complementarity in Human--AI teams
requires deliberate design: simply adding a human or an AI to the loop does
not guarantee better outcomes~\citep{Kamar2016,Bansal2021}. Recent work argues
that complementarity depends on explicit role partitioning, trust
calibration, shared mental models, and escalation structures rather than on
nominal human presence alone~\citep{Gonzalez2026}, and the
meta-analysis by \citet{Vaccaro2024} confirms that Human--AI combinations do
\emph{not} reliably outperform the better standalone agent, especially in
decision-making tasks.
For allocation research, the implication is direct: the benefit of Human--AI
collaboration depends on how work is decomposed and governed, not merely on
adding a human review step to an AI pipeline.
Field experiments with large language models show that GenAI substantially
raises productivity in writing~\citep{Noy2023} and consulting
tasks~\citep{DellAcqua2023}, and reduces task completion time for software
developers by over 50\%~\citep{Peng2023}.
Crucially, \citet{DellAcqua2023} report a \emph{jagged frontier}: AI
outperforms humans on some tasks while underperforming on others, reinforcing
the need for dynamic, task-level allocation rather than blanket automation.
\citet{Raisch2021} frame this as the automation--augmentation paradox:
the same AI capability that increases efficiency may simultaneously erode the
human expertise required to govern, audit, and correct it.
\textsc{HAAS} translates this insight into practice by treating the jagged
frontier as a design parameter rather than a fixed obstacle: it provides a
systematic, pre-deployment way to identify which tasks belong on each side of
the frontier under a given governance contract.

Translating these principles to autonomous Cyber-Physical Systems (CPS)
requires explicit design of the human integration itself.
\citet{Gil2019} characterise the technological challenges of integrating
humans into the CPS autonomy loop and propose a conceptual framework ---
validated on an autonomous vehicle prototype --- for specifying how humans
and autonomous systems should cooperate, making explicit role assignments
and participation boundaries that HAAS extends to the task-allocation level.
In human--robot collaboration, adaptive task-allocation frameworks have moved
beyond static role assignment: recent work models trust~\citep{Ali2022},
adapts assignments via augmented reality and digital twins~\citep{Petzoldt2025},
responds to operator fatigue or stress~\citep{Urrea2025,KirgilBudakli2025},
and incorporates worker preferences via reinforcement learning~\citep{Wang2025}.
\Cref{tab:positioning} positions \textsc{HAAS} relative to these studies;
the key differences are the explicit five-mode autonomy vocabulary, the
separate governance layer, and the cross-domain reproducible benchmark.

The broader paradigm motivating this work is \emph{Hybrid Intelligence}:
the design of Human--AI systems in which humans and AI complement each other
so that the team achieves outcomes neither could reach independently, while
humans retain authority over goals and values~\citep{Dellermann2019}.
\textsc{HAAS} operationalises this paradigm at the task-allocation level:
rather than choosing a single automation level for a workflow, it governs the
conditions under which each collaboration mode is appropriate, ensuring that
efficiency gains and human capability are co-optimised rather than traded off.

Despite this progress, most allocation proposals still optimise a single
workflow or local productivity/ergonomics objective.
They rarely provide all of the following together:
(\emph{i})~an explicit rule layer for governance;
(\emph{ii})~a graded autonomy vocabulary beyond binary assignment;
and (\emph{iii})~a reproducible cross-domain benchmark for pre-deployment
policy analysis.
This combined gap is what \textsc{HAAS} addresses.

\subsection{Learning-Based Allocation Strategies}
\label{sec:bg:learning}

A policy-aware allocation engine must not only encode domain knowledge
but also \emph{adapt} to outcome feedback as the task mix and human state
evolve over time.
This requirement points to sequential decision-making methods that balance
exploration of uncertain allocation options with exploitation of accumulated
experience.

Sequential decision-making under uncertainty has been formalised through
the multi-armed bandit (MAB) paradigm~\citep{Lattimore2020}.
UCB-based algorithms~\citep{Auer2002} balance exploration and exploitation
through optimistic upper confidence bounds on estimated arm rewards.
LinUCB~\citep{Li2010} extends this to contextual settings where side
information (here, the cognitive dimensions of a subtask) is available,
making it a natural fit for task-heterogeneous workflows.
Thompson Sampling~\citep{Russo2018} offers a Bayesian alternative that has
shown competitive regret in practice.

A contextual-bandit formulation is appropriate here because each allocation
decision is made at the subtask level and receives immediate evaluative
feedback after execution, while governance rules externally restrict the
feasible action set. Supervised learning would require labelled ``optimal''
allocations that are not available ex ante, and a full reinforcement-learning
formulation would introduce state-transition modelling and long-horizon credit
assignment beyond what this benchmark is designed to estimate. We therefore
adopt the simplest online-learning model that can exploit task context, learn
from partial feedback, and remain auditable under explicit policy constraints.

The discounted-UCB variant~\citep{Garivier2011} addresses non-stationary
environments by exponentially down-weighting historical rewards --- a natural
fit for workflows where task mix and human state evolve over time.
Applications of bandits to workload management and scheduling exist in
manufacturing~\citep{Gijsbrechts2022} and sequential resource allocation
more broadly~\citep{Lattimore2020}, but rarely
incorporate human well-being metrics or governance constraints as part of
the reward signal.
\textsc{HAAS} treats fatigue and deskilling as first-class outcomes alongside
efficiency and quality, embedding them directly in the bandit reward signal.

\subsection{Governance and Responsible AI in Work Systems}
\label{sec:bg:governance}

Even a technically optimal allocation policy is insufficient for deployment
if it operates without organisational constraints.
Responsible AI deployment in work systems requires mechanisms that preserve
human oversight, calibrate trust appropriately, and comply with regulatory
and safety obligations.
This subsection reviews the governance and responsible AI literature that
motivates the \texttt{PolicyEngine} component of \textsc{HAAS}.

A recurrent idea in responsible AI is that HITL arrangements should preserve
active human involvement in the workflow, especially in high-stakes settings
~\citep{Monarch2021}. But keeping HITL is not enough by itself.
What also matters is \emph{appropriate reliance}~\citep{Lee2004}: people
should neither over-trust nor unnecessarily reject AI support.
\citet{Bucinca2021} show empirically that people over-rely on AI
recommendations even when those recommendations are wrong, and that cognitive
forcing interventions can partially mitigate this bias.
Regulatory frameworks such as the EU AI Act~\citep{EUAIAct2024} impose
transparency and human oversight obligations for high-risk AI applications;
collaborative robotics standards~\citep{ISO10218,ISO15066} specify authority
boundaries for shared workspaces; and human-centred AI guidelines
\citep{Amershi2019,Shneiderman2022} codify principles for effective human
oversight of AI decisions. In parallel, explainability research emphasises
that useful explanations are social and context-dependent rather than mere
exposure of internal model logic, which matters directly for calibrated
reliance in AI-supported decisions~\citep{Miller2019}.
One systematic review of adaptive autonomy synthesises a decade of human
factors research and shows that the literature remains fragmented with respect
to adaptation triggers, evaluation criteria, and operational design
choices~\citep{Hauptman2024}.

The adaptive automation literature~\citep{Inagaki2003,Kaber2004} offers an
important precedent: systems that dynamically shift automation levels in
response to operator state (workload, fatigue, situation awareness) have been
studied since the 1990s.
\textsc{HAAS} extends this idea to LLM-era AI capabilities and explicit
organisational governance.
Existing HITL approaches, however, are predominantly architectural --- they
specify \emph{where} humans should be in the loop but not \emph{how} the
loop should adapt dynamically to task mix, fatigue, or strategic priorities.
Our \texttt{PolicyEngine} operationalises governance as a set of evaluable
rules that constrain, but do not replace, the learning-based allocator.
To our knowledge, \textsc{HAAS} is among the first frameworks to jointly provide:
(\emph{i})~a graded five-mode autonomy vocabulary for knowledge-work
allocation; (\emph{ii})~an editable rule-based governance layer that is
structurally prior to the adaptive learner; (\emph{iii})~explicit human-state
dynamics (fatigue, trust, deskilling) embedded directly in the reward signal;
and (\emph{iv})~a reproducible cross-domain benchmark spanning software
engineering and manufacturing.

\begin{table}[t]
  \centering
  \caption{Comparison of \textsc{HAAS} with representative allocation studies
    along the dimensions most relevant to this paper.}
  \label{tab:positioning}
  \footnotesize
  \setlength{\tabcolsep}{2pt}
  \renewcommand{\arraystretch}{1.08}
  \begin{adjustbox}{max width=\linewidth}
  \rowcolors{2}{gray!8}{}
  \begin{tabular}{@{}lcccccc@{}}
    \toprule
    \textbf{Study} &
    \textbf{\shortstack{Adaptive\\alloc.}} &
    \textbf{\shortstack{Autonomy\\model}} &
    \textbf{\shortstack{Governance\\layer}} &
    \textbf{\shortstack{Human factors\\modelled}} &
    \textbf{\shortstack{Repro-\\ducible}} &
    \textbf{\shortstack{Evaluation\\scope}} \\
    \midrule
    \citet{Ali2022}            & yes & not explicit & \shortstack{no separate\\layer} & trust & no & \shortstack{single HRC\\team} \\[3pt]
    \citet{Petzoldt2025}       & yes & \shortstack{workflow-\\specific} & not explicit & \shortstack{worker\\performance} & no & single assembly \\[3pt]
    \citet{Urrea2025}          & yes & not explicit & not explicit & fatigue + skill & no & single HRC sim. \\[3pt]
    \citet{Wang2025}           & yes & not explicit & not explicit & worker preference & no & single assembly \\[3pt]
    \textsc{HAAS} (this paper) & yes & five modes & \shortstack{explicit\\rule layer} & \shortstack{fatigue + trust\\+ deskilling} & \textbf{yes} & \shortstack{cross-domain\\benchmark} \\
    \bottomrule
  \end{tabular}%
  \end{adjustbox}
  \par\vspace{2pt}
  \parbox{0.96\linewidth}{\raggedright\footnotesize
    ``Governance layer'' refers to a distinct, editable rule layer rather
    than implicit safety logic or embodied constraints.
    ``Reproducible'' indicates fixed seeds, published parameter tables, and a
    command-line benchmark runner.}
\end{table}

\Cref{tab:positioning} makes the positioning claim explicit.
Prior studies treat allocation as adaptive but typically operate within a
single embodied workflow, encode autonomy only implicitly, lack a separate
governance layer, and do not provide reproducible multi-domain benchmarks.
\textsc{HAAS} differs on all four dimensions.

%% ================================================================
\section{The HAAS Framework}
\label{sec:framework}
%% ================================================================

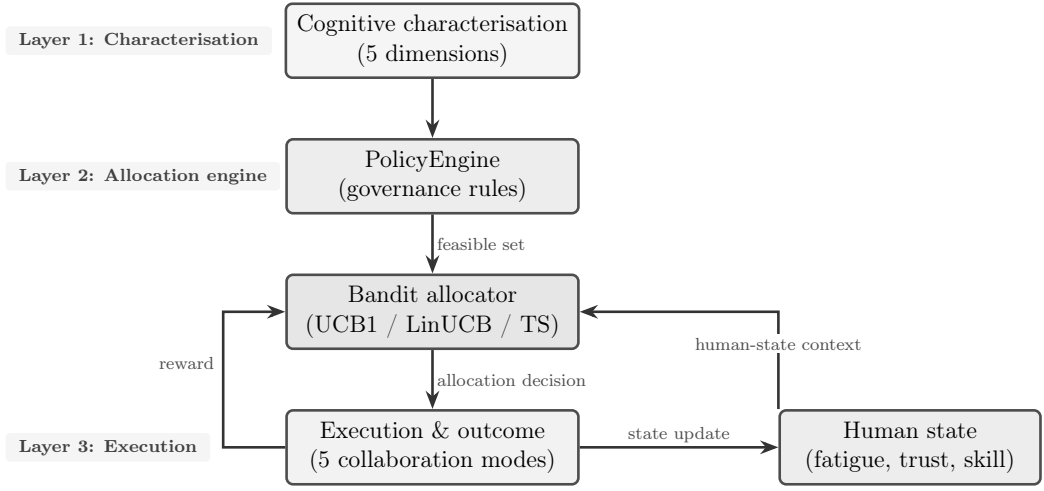
\begin{figure}[htbp]
  \centering
  \resizebox{\linewidth}{!}{%
    \begin{tikzpicture}[
      box/.style={draw=black!70, very thick, rounded corners=3pt,
                  minimum width=4.7cm, minimum height=1.0cm,
                  align=center, font=\small, inner sep=5pt},
      sidebox/.style={box, minimum width=4.2cm},
      layer/.style={font=\scriptsize\bfseries, text=black!75,
                    fill=black!4, rounded corners=2pt,
                    inner xsep=6pt, inner ysep=3pt},
      arr/.style={-Stealth, very thick, draw=black!80},
      lab/.style={font=\scriptsize, text=black!70, fill=white, inner sep=1.5pt}
    ]
    \node[box, fill=black!4] (cog) {Cognitive characterisation\\(5 dimensions)};
    \node[box, fill=black!7, below=0.95cm of cog] (pol) {PolicyEngine\\(governance rules)};
    \node[box, fill=black!10, below=0.95cm of pol] (ban) {Bandit allocator\\(UCB1 / LinUCB / TS)};
    \node[box, fill=black!5, below=0.95cm of ban] (exe) {Execution \& outcome\\(5 collaboration modes)};
    \node[sidebox, fill=black!7, right=3.2cm of exe] (hs) {Human state\\(fatigue, trust, skill)};

    \coordinate (labelx) at ([xshift=-4.5cm]cog.west);
    \node[layer, anchor=west] at (labelx |- cog) {Layer 1: Characterisation};
    \node[layer, anchor=west] at (labelx |- pol) {Layer 2: Allocation engine};
    \node[layer, anchor=west] at (labelx |- exe) {Layer 3: Execution};

    \draw[arr] (cog) -- (pol);
    \draw[arr] (pol) -- node[right, lab] {feasible set} (ban);
    \draw[arr] (ban) -- node[right, lab] {allocation decision} (exe);
    \draw[arr] (exe) -- node[above, lab] {state update} (hs);
    \draw[arr] (hs.north west) |- node[pos=0.28, above, lab] {human-state context} (ban.east);
    \draw[arr] (exe.west) -- ++(-1.0,0) coordinate (rewardturn)
      -- node[pos=0.62, left=2pt, lab] {reward} (rewardturn |- ban.west)
      -- (ban.west);
    \end{tikzpicture}%
  }
  \caption{Three-layer architecture of \textsc{HAAS}. The governance layer
           filters feasible collaboration modes before the bandit selects an
           allocation; execution outcomes then update both reward and human
           state for the next cycle.}
  \label{fig:architecture}
\end{figure}

\textsc{HAAS} is organised into three layers (\Cref{fig:architecture}).
Layer~1 scores each subtask on five cognitive dimensions.
Layer~2 is the allocation engine: the \texttt{PolicyEngine} applies governance
constraints first, then the bandit selects among the remaining feasible modes.
Layer~3 maps the decision to one of five collaboration modes, tracks outcomes,
and updates human state, feeding back into the next cycle.
The key design principle is that governance is applied \emph{before} the
bandit observes the reward signal, so learned behaviour always operates within
organisationally acceptable bounds.
Throughout the framework and experiments, the atomic allocation unit is the
\emph{subtask}; we retain \emph{task} as a generic umbrella term and
\emph{task type} for the class label attached to a subtask in the benchmark
catalogue.
The remainder of this section details each component in turn: cognitive
characterisation and affinity scoring
(\Cref{sec:framework:cognitive,sec:framework:weights}), collaboration modes
(\Cref{sec:framework:modes}), the allocation engine
(\Cref{sec:framework:engine}), the execution loop
(\Cref{sec:framework:execution}), and the human-state model
(\Cref{sec:framework:human}).

\subsection{Cognitive Characterisation of Subtasks}
\label{sec:framework:cognitive}

Each subtask must be characterised in a way that makes its suitability for
human or AI execution explicit and computable.
\textsc{HAAS} uses a five-dimension scoring instrument grounded in cognitive
task analysis~\citep{Crandall2006} and the four-stage automation taxonomy of
\citet{Parasuraman2000}, translating qualitative task properties into a
continuous AI affinity score.

Each subtask $s$ is characterised by a vector
$\mathbf{d}(s) = (r, \tau, c, a, h) \in [0,1]^5$ of rubric-based scores
assigned on a common normalised 0--1 scale, where:

\begin{itemize}
  \item $r$ --- \textbf{repetitiveness}: the degree to which the subtask is
    mechanical, templated, or rule-governed;
  \item $\tau$ --- \textbf{technical depth}: the level of domain expertise
    required for competent execution;
  \item $c$ --- \textbf{creativity}: the degree to which the subtask demands
    novel problem-solving or aesthetic judgement;
  \item $a$ --- \textbf{ambiguity}: the extent to which requirements or success
    criteria are underspecified;
  \item $h$ --- \textbf{human interaction}: the degree to which the subtask
    requires interpersonal communication, negotiation, or empathy.
\end{itemize}

From these five dimensions we derive a scalar \emph{AI affinity}:

\begin{equation}
  \alpha_{\mathrm{AI}}(s) =
    w_r \, r + w_\tau \, \tau
    + w_c \,(1 - c)
    + w_a \,(1 - a)
    + w_h \,(1 - h)
  \label{eq:affinity}
\end{equation}

where $\mathbf{w} = (w_r, w_\tau, w_c, w_a, w_h)$ is a non-negative weight
vector summing to unity. Human affinity is
$\alpha_{\mathrm{H}}(s) = 1 - \alpha_{\mathrm{AI}}(s)$. The rationale for these weights,
and the extent to which the affinity score changes under alternative weight
choices, are discussed in \Cref{sec:framework:weights}.

\begin{table}[!t]
  \centering
  \caption{Representative software (SW) and manufacturing (MF) subtasks,
    ordered by AI affinity $\alpha_{\mathrm{AI}}$ (\Cref{eq:affinity}).}
  \label{tab:subtask_examples}
  \footnotesize
  \setlength{\tabcolsep}{3.5pt}
  \begin{adjustbox}{max width=\linewidth}
  \begin{tabular}{@{}l>{\raggedright\arraybackslash}p{2.6cm}>{\raggedright\arraybackslash}p{2.3cm}cccccrl@{}}
    \toprule
    Dom. & Subtask & Task Type & $r$ & $\tau$ & $c$ & $a$ & $h$ &
    $\alpha_{\mathrm{AI}}$ & Constraint \\
    \midrule
    \multicolumn{10}{l}{\textit{Human-centric\; ($\alpha_{\mathrm{AI}} < 0.45$)}} \\[2pt]
    SW & Stakeholder Interview    & Req.\ Analysis  & 0.15 & 0.25 & 0.55 & 0.80 & 0.95 & 0.23 & --- \\
    MF & Safety Incident Mgmt.   & Safety Superv.  & 0.05 & 0.55 & 0.65 & 0.90 & 0.95 & 0.24 & \textbf{Human-Only} \\
    SW & Root-Cause Analysis      & Debugging       & 0.10 & 0.85 & 0.70 & 0.65 & 0.50 & 0.39 & --- \\
    MF & Defect Analysis          & Quality Insp.   & 0.20 & 0.70 & 0.60 & 0.70 & 0.55 & 0.40 & --- \\
    \midrule
    \multicolumn{10}{l}{\textit{Balanced\; ($0.45 \leq \alpha_{\mathrm{AI}} < 0.70$)}} \\[2pt]
    SW & Business-Logic Coding    & Code Gen.       & 0.30 & 0.70 & 0.60 & 0.50 & 0.40 & 0.47 & --- \\
    SW & API Design               & Arch.\ Design   & 0.30 & 0.70 & 0.55 & 0.40 & 0.45 & 0.48 & --- \\
    MF & Sensor Data Analysis     & Pred.\ Maint.   & 0.45 & 0.75 & 0.40 & 0.50 & 0.20 & 0.59 & --- \\
    MF & Precision Assembly       & Assembly        & 0.65 & 0.55 & 0.25 & 0.22 & 0.22 & 0.67 & --- \\
    SW & API Documentation        & Documentation   & 0.70 & 0.40 & 0.15 & 0.15 & 0.10 & 0.69 & --- \\
    \midrule
    \multicolumn{10}{l}{\textit{AI-centric\; ($\alpha_{\mathrm{AI}} \geq 0.70$)}} \\[2pt]
    MF & Visual Inspection        & Quality Insp.   & 0.75 & 0.50 & 0.10 & 0.20 & 0.15 & 0.73 & --- \\
    MF & AGV Route Mgmt.$^\dagger$ & Logistics       & 0.82 & 0.45 & 0.12 & 0.15 & 0.18 & 0.74 & \textbf{AI-Only} \\
    SW & Boilerplate Generation   & Code Gen.       & 0.90 & 0.30 & 0.10 & 0.10 & 0.05 & 0.76 & --- \\
    MF & Production Cycle         & Machine Op.     & 0.92 & 0.30 & 0.05 & 0.08 & 0.15 & 0.76 & \textbf{AI-Only} \\
    \bottomrule
    \multicolumn{10}{@{}p{\linewidth}@{}}{\footnotesize
      $^\dagger$AGV: Automated Guided Vehicle.} \\
  \end{tabular}
  \end{adjustbox}
\end{table}

The five dimensions form a compact, auditable instrument: every allocation
decision can be traced back to the dimension scores that drove it, supporting
the transparency requirements of governance-oriented deployments.

\Cref{tab:subtask_examples} shows representative subtasks from the
50-task benchmark (25 per domain), ordered by increasing AI affinity.
Dimension scores are reported in $[0,1]$, with
$r$=repetitiveness, $\tau$=technical depth, $c$=creativity,
$a$=ambiguity, and $h$=human interaction.
\emph{Human-Only} and \emph{AI-Only} constraints denote hard policy rules
enforced by the \texttt{PolicyEngine} independently of the learned allocator.

\subsection{Affinity Weight Specification}
\label{sec:framework:weights}

The default weight vector $\mathbf{w} = (0.35,\;0.25,\;0.20,\;0.10,\;0.10)$
is grounded in the human--automation literature.
\textbf{Repetitiveness} receives the largest weight ($w_r = 0.35$) because
rule-governed work is the canonical automation target~\citep{Parasuraman2000}
and yields the largest AI productivity gains~\citep{Peng2023}.
\textbf{Technical depth} ($w_\tau = 0.25$) is positive because contemporary AI
performs strongly on well-structured technical tasks inside the technological
frontier~\citep{Brynjolfsson2023,DellAcqua2023}, though organisations in
safety-critical domains should treat this as a starting point.
\textbf{Creativity} ($w_c = 0.20$, inverted), together with
\textbf{ambiguity} and \textbf{human interaction}
($w_a = w_h = 0.10$, both inverted), is weighted this way to reflect
jagged-frontier evidence that open-ended and interpersonal tasks still tend
to favour human performance~\citep{DellAcqua2023,Noy2023,McNeese2018}.

\paragraph{Sensitivity check}
Perturbing each weight by $\pm 30\%$ (re-normalised) preserved the
rank-ordering of all strategies and the direction of all governance effects
(\Cref{sec:results:sensitivity}).

\subsection{The Autonomy Spectrum: Five Collaboration Modes}
\label{sec:framework:modes}

With each subtask characterised, the next design question is what an
``allocation'' actually means in practice.
\textsc{HAAS} resolves this through a five-mode
\emph{autonomy spectrum} $\mathcal{M}$ (see \Cref{tab:modes}),
extending prior taxonomies of automation levels~\citep{Sheridan1992} and
robot autonomy~\citep{Beer2014} to the specific demands of knowledge-work
allocation.
Each mode $m \in \mathcal{M}$ is characterised by a human participation
share $\sigma_{\mathrm{H}}(m) \in [0,1]$ and an AI participation share
$\sigma_{\mathrm{AI}}(m) = 1 - \sigma_{\mathrm{H}}(m)$.

\begin{table}[H]
  \centering
  \caption{The five collaboration modes of the autonomy spectrum.}
  \label{tab:modes}
  \footnotesize
  \begin{adjustbox}{max width=\linewidth}
  \begin{tabular}{@{}>{\raggedright\arraybackslash}p{2.35cm}cc>{\raggedright\arraybackslash}p{6.5cm}@{}}
    \toprule
    \textbf{Mode} & $\sigma_{\mathrm{H}}$ & $\sigma_{\mathrm{AI}}$ &
    \textbf{Semantics} \\
    \midrule
    \textsc{Human-Only}  & 1.0 & 0.0 &
      Human executes alone; AI not involved. \\
    \textsc{Copilot}     & $>0.5$ & $<0.5$ &
      Human leads; AI assists in real time. \\
    \textsc{Peer}        & $\approx0.5$ & $\approx0.5$ &
      Complementary parallel execution; neither leads. \\
    \textsc{Supervised}  & $<0.5$ & $>0.5$ &
      AI leads; human validates before applying. \\
    \textsc{Autonomous}  & 0.0 & 1.0 &
      AI executes alone; human not involved. \\
    \bottomrule
  \end{tabular}
  \end{adjustbox}
\end{table}

Modes with $\sigma_{\mathrm{H}} > 0$ and $\sigma_{\mathrm{AI}} > 0$ are
\emph{shared} modes (\textsc{Copilot}, \textsc{Peer}, \textsc{Supervised}).
The \textsc{Peer} mode is selected when the difference between human and AI
shares falls within a balance threshold $\delta_{\mathrm{peer}}$ (default
$0.20$); otherwise the dominant partner determines whether the mode is
\textsc{Copilot} (human primary) or \textsc{Supervised} (AI primary).
In the \textsc{Peer} mode, shares are held at the equal-split nominal
values ($\sigma_{\mathrm{H}} = \sigma_{\mathrm{AI}} = 0.5$) without the
fatigue-driven dynamic adjustment applied in
\Cref{eq:copilot_share,eq:supervised_share}; the balance threshold
$\delta_{\mathrm{peer}}$ governs mode entry but not within-mode
redistribution.

The values in \Cref{tab:modes} are \emph{baseline} shares that define the
nominal character of each mode. In the three shared modes, these shares are
further adjusted at execution time as functions of the current human fatigue
$f \in [0,1]$ and subtask complexity, producing \emph{dynamic} shares
$\sigma_{\mathrm{AI}}^{\mathrm{Copilot}}(s,f)$ and
$\sigma_{\mathrm{H}}^{\mathrm{Supervised}}(s,f)$ (mode superscript, arguments in
parentheses). The mode-selection label remains fixed by the policy; only
the within-mode split adapts:

\begin{align}
  \sigma_{\mathrm{AI}}^{\mathrm{Copilot}}(s,f)
    &= \mathrm{clip}\!\left(
        0.60\,p_f + 0.20\,\kappa(s),\;
        \sigma_{\min},\; \sigma_{\max}
       \right)
  \label{eq:copilot_share} \\[4pt]
  \sigma_{\mathrm{H}}^{\mathrm{Supervised}}(s,f)
    &= \mathrm{clip}\!\left(
        0.35\,j(s) - 0.50\,\max(0,\, f - 0.65),\;
        \sigma_{\min}^H,\; \sigma_{\max}^H
       \right)
  \label{eq:supervised_share}
\end{align}

where $p_f = \max(0, f - 0.35)$ is the excess fatigue above the hybrid
trigger threshold, $\kappa(s) = (\tau + a)/2$ is a subtask complexity proxy,
and $j(s) = (a + h)/2$ is a judgment-need proxy.
Here, $\mathrm{clip}(x, a, b)$ truncates $x$ to the admissible interval
$[a,b]$: values below $a$ are set to $a$, and values above $b$ are set to $b$.
Bounds are $(\sigma_{\min}, \sigma_{\max}) = (0.20, 0.55)$ for Copilot AI
share and $(0.10, 0.40)$ for Supervised human share.
The Copilot upper bound of $0.55$ allows the within-mode AI share to
transiently exceed $0.5$ under high fatigue or complexity, which may appear
inconsistent with the $\sigma_{\mathrm{H}} > 0.5$ condition in
\Cref{tab:modes}. The resolution is that the mode label is fixed by the
governance policy at \emph{mode-selection time}, not by the real-valued
participation share at execution time. The $\sigma_{\mathrm{H}} > 0.5$
entry in \Cref{tab:modes} therefore describes the nominal character of
\textsc{Copilot} --- the human leads the subtask and retains authority --- not
a strict runtime invariant of the share equation. In design terms, a human
can remain the accountable lead even when the AI contributes the larger
share of mechanical execution, which is the intended semantics of the
\textsc{Copilot} mode.

\noindent\textit{Example.} For a mid-complexity subtask ($\kappa(s){=}0.55$,
$j(s){=}0.45$) at fatigue $f{=}0.50$ (so $p_f{=}0.15$), \Cref{eq:copilot_share}
gives $\sigma_{\mathrm{AI}}^{\mathrm{Copilot}} = \mathrm{clip}(0.60{\times}0.15 +
0.20{\times}0.55,\,0.20,\,0.55) = \mathrm{clip}(0.20,\ldots) = 0.20$, leaving
the human with an 80\% share. At $f{=}0.80$ ($p_f{=}0.45$), the same subtask
yields $\sigma_{\mathrm{AI}}^{\mathrm{Copilot}} = \mathrm{clip}(0.38,\ldots) = 0.38$,
illustrating how higher fatigue shifts more work toward the AI assistant.

\subsection{Policy-Aware Allocation Engine}
\label{sec:framework:engine}

With the cognitive characterisation and the autonomy spectrum defined,
the central question becomes: how does \textsc{HAAS} decide which mode
to apply to a given subtask at a given moment?
This decision must satisfy two competing requirements simultaneously:
it must respect hard organisational constraints regardless of what the
data suggest, and it must adapt to accumulated outcome evidence to improve
allocation quality over time.

The allocation engine operates in two sequential stages: governance
pre-filtering and bandit-based learning (see \Cref{alg:allocation}).

\paragraph{Stage 1 --- Governance pre-filtering (PolicyEngine)}
A \texttt{PolicyEngine} evaluates an ordered list of governance rules
against the current subtask $s$, operating as a forward-chaining
production rule system~\citep{Turban1992,Giarratano1994,Liao2005}: rules fire in
priority order (lower values take precedence), and the first matching rule
wins, producing a deterministic governance directive.
Each rule specifies the conditions under which it applies (task type, subtask
name, human-state threshold) and what it enforces: a designated lead agent
(\textsc{Human} or \textsc{AI}), a mandatory collaboration mode, or an
autonomy cap that limits how far AI execution may proceed.
The resulting directive therefore does one of two things: it either fixes the
assignment outright, or it restricts the admissible portion of the autonomy
spectrum within which learning may operate.
If no rule matches, the engine applies the system-wide default cap, leaving
the bandit free to choose anywhere within the permitted spectrum.
This design ensures that organisational governance is structurally prior to
learned behaviour, not a post-hoc filter.

\paragraph{Stage 2 --- Contextual bandit learning}
If no policy forces the assignment, a bandit allocator selects a
collaboration mode from the feasible set permitted by the active governance
constraints.
The bandit therefore maintains one arm per collaboration mode
(\textsc{Human-Only}, \textsc{Copilot}, \textsc{Peer}, \textsc{Supervised},
\textsc{Autonomous}) for each task type.
The framework supports four algorithms. However, the empirical benchmark
reported here focuses on UCB1, LinUCB, and Thompson Sampling, and leaves
Discounted-UCB out because the 8-sprint horizon is too short for reward
discounting to produce behaviour meaningfully different from UCB1:

\begin{itemize}
  \item \textbf{UCB1}~\citep{Auer2002}: optimistic upper confidence bound
    $Q_k + C\sqrt{\ln N / n_k}$ where $C$ is an exploration constant.
  \item \textbf{Discounted-UCB} (D-UCB): UCB1 with exponential discount
    $\gamma^t$; implemented for longer or more clearly non-stationary
    deployments, but not analysed further here because with $\gamma{=}0.97$
    and only 8 sprints it behaves almost identically to UCB1.
  \item \textbf{LinUCB}~\citep{Li2010}: linear contextual bandit that
    uses $\mathbf{d}(s)$ as the context vector; reward is modelled as
    $\mathbf{d}(s)^\top \boldsymbol{\theta}_k + \epsilon$.
  \item \textbf{Thompson Sampling}~\citep{Russo2018}: samples $\tilde{Q}_k$
    from a Beta posterior, selecting $\arg\max_k \tilde{Q}_k$.
\end{itemize}

During the first $T_{\mathrm{explore}}$ sprints (set to 3 for the 8-sprint
benchmark; see \Cref{sec:setup:strategies}), a heuristic warm-start
initialises the bandit over the feasible modes before online learning begins.
The framework can also be seeded from a human managerial configuration
specified through the setup workflow; in the benchmark reported here,
however, all learned allocators use the same heuristic warm-start so that
their online adaptation is compared from a common initial condition.

\paragraph{Reward signal}
After execution the bandit receives a scalar reward, whose exact form depends
on the active reward profile. The main benchmark in this paper uses a
\emph{four-outcome} reward that combines quality, time, cost, and well-being:
\begin{equation}
  R = w_q \cdot q + w_t \cdot (1 - \bar{t}) + w_{\mathrm{co}} \cdot (1 - \bar{c}) + w_w \cdot W
  \label{eq:reward}
\end{equation}
where $q$ is subtask quality, $\bar{t}$ is normalised time taken, $\bar{c}$ is
normalised cost, and $W$ is a well-being composite that penalises fatigue,
monotony, deskilling, and exclusion while rewarding fatigue relief (formal
definition of $W$ given in \Cref{sec:setup:platform}).
The main-table benchmark uses $(w_q, w_t, w_{\mathrm{co}}, w_w) = (0.30, 0.20, 0.10, 0.40)$.
Cumulative allocation regret --- the non-negative reward gap relative to the
better of the two pure counterfactual baselines --- is defined formally in
\Cref{sec:setup:metrics} and serves as the primary learning diagnostic.
Additional supported reward profiles are \emph{efficiency}, \emph{quality-first},
and \emph{symbiosis}; these are predefined presets within a more general
configuration space in which users may set arbitrary weight combinations over
quality, time, cost, and well-being to express narrower or mixed decision
objectives.
The learner therefore chooses directly among the feasible collaboration modes.
Once a mode has been selected, the system derives the corresponding lead agent
and the human/AI participation shares at execution time. This keeps the
learning problem simple while preserving the richer collaboration structure
captured by the framework.

\Cref{alg:allocation} makes the decision logic explicit. The sequence is
simple: governance is checked first, then learning operates only within the
set of modes that governance permits. If a policy rule forces a mode, that
decision is applied directly. Otherwise, the policy layer defines the feasible
set through the active autonomy cap, and the bandit selects within that set.
The chosen mode is finally translated into the allocation specification
$(m,\sigma_{\mathrm{H}},\sigma_{\mathrm{AI}})$.

\begin{algorithm}[ht]
  \caption{HAAS Allocation Decision}
  \label{alg:allocation}
  \begin{algorithmic}[1]
    \Require subtask $s$, human state $(f, \mathit{trust}, \mathit{skill})$,
             policy rules $\mathcal{P}$, bandit $\mathcal{B}$
    \Ensure allocation specification $(m, \sigma_{\mathrm{H}}, \sigma_{\mathrm{AI}})$
    \State $d_{\mathrm{gov}} \leftarrow \textsc{PolicyEngine}(\mathcal{P}, s)$ \Comment{governance directive}
    \If{$d_{\mathrm{gov}}$ prescribes a direct assignment}
      \State \textbf{return} assignment implied by $d_{\mathrm{gov}}$
    \Else
      \State $\mathcal{M}_{\mathrm{feas}} \leftarrow \text{modes admissible under } d_{\mathrm{gov}}$
      \State $m \leftarrow \textsc{BanditSelect}(\mathcal{B}, s, f, \mathcal{M}_{\mathrm{feas}})$
      \State \textbf{return} \textsc{InstantiateMode}(m, s, f)
    \EndIf
  \end{algorithmic}
\end{algorithm}

Representative rules help illustrate how this works in practice.
Regulatory-compliance subtasks may force \textsc{Human-Only}; onboarding periods
may force \textsc{Copilot}; and critical fatigue ($f > 0.90$) may force
\textsc{Autonomous}. When no rule forces a specific mode, the policy layer
still constrains the choice through the active autonomy cap. For example, if
the cap is \textsc{Supervised}, the bandit may select any mode except
\textsc{Autonomous}.
In the IF-THEN notation of classical production-rule systems~\citep{Giarratano1994}, representative rules are:

\begin{center}
{\ttfamily\footnotesize
\begin{tabularx}{0.96\linewidth}{@{}l>{\raggedright\arraybackslash}X>{\raggedright\arraybackslash}p{0.33\linewidth}@{}}
R1: & IF task.type = \textit{regulatory-review} &
THEN force \textsc{Human-Only}\\[3pt]
R2: & IF sprint $\leq 2\;\wedge\;$task.type = \textit{architecture} &
THEN force \textsc{Copilot},\ cap\,=\,\textsc{Supervised}\\[3pt]
R3: & IF fatigue $> 0.90$ &
THEN force \textsc{Autonomous}\ \textnormal{\small[safety exception: operator incapacitation; not a general governance preference]}\\[3pt]
R4: & IF (no match) &
THEN cap\,=\,system-wide default
\end{tabularx}}
\end{center}

\noindent Rules fire in priority order; the first match wins and short-circuits
the remaining rules, matching the standard conflict-resolution strategy of
forward-chaining expert systems.

\Cref{tab:policy_catalogue} lists the complete benchmark rule catalogue,
showing how the three structurally distinct rule types
(hard assignment, mode-forcing, and autonomy-cap) are distributed across
the two domains and the governance levels at which they activate.
Governance levels L1--L4 define cumulative layers of constraint severity,
as summarised in \Cref{tab:gov_ladder_config}.

\begin{table}[H]
  \centering
  \caption{Governance Ladder benchmark levels.}
  \label{tab:gov_ladder_config}
  \footnotesize
  \setlength{\tabcolsep}{3.5pt}
  \renewcommand{\arraystretch}{1.08}
  \begin{adjustbox}{max width=\linewidth}
  \begin{tabular}{@{}llccp{0.47\linewidth}@{}}
    \toprule
    \textbf{Level} & \textbf{Label} &
    \textbf{\shortstack{Risk\\threshold}} &
    \textbf{\shortstack{Autonomy budget\\threshold}} &
    \textbf{Added constraints} \\
    \midrule
    L0 & None        & ---  & ---  &
      Policy layer disabled; allocator unconstrained. \\
    L1 & Light       & 0.72 & 0.45 &
      No task-specific rules; only a permissive autonomy cap. \\
    L2 & Moderate    & 0.62 & 0.58 &
      No task-specific rules; a stricter autonomy cap. This is the calibrated
      baseline used in the strategy comparison tables. \\
    L3 & Strong      & 0.52 & 0.68 &
      Adds supervised execution on selected high-impact subtask types in both
      domains, on top of the autonomy cap. \\
    L4 & Very strong & 0.40 & 0.78 &
      Adds human-only execution on the most critical subtask types and supervised
      execution on additional subtask types, producing a compliance-first regime. \\
    \bottomrule
  \end{tabular}
  \end{adjustbox}
  \par\smallskip
  \begin{minipage}{\linewidth}
    \footnotesize\noindent
    Detailed rule sets by domain are reported in \Cref{tab:policy_catalogue}.
    Rules fire in priority order (lower = higher precedence); the first
    match short-circuits remaining rules. In L1--L4, the experience-based
    autonomy cap remains active in addition to any task-specific rules.
  \end{minipage}
\end{table}

The full policy engine for a given level is assembled by composing the rules
active at that level with the experience-based autonomy cap
(\Cref{tab:gov_ladder_config}), so the effective constraint at any given
subtask is the strictest applicable rule plus the cap, evaluated in
priority order.

\begin{table}[H]
  \centering
  \caption{Benchmark \texttt{PolicyEngine} rule catalogue.
    ``Hard'' rules force a pure allocation outcome regardless of the
    bandit. ``Mode'' rules force a collaboration mode while leaving
    within-mode participation shares to the framework. The autonomy cap ($\star$) is a parametric mechanism
    active at all L1--L4 levels; its thresholds vary by level
    (see \Cref{tab:gov_ladder_config}).}
  \label{tab:policy_catalogue}
  \footnotesize
  \setlength{\tabcolsep}{3.5pt}
  \renewcommand{\arraystretch}{1.08}
  \begin{adjustbox}{max width=\linewidth}
  \begin{tabular}{@{}lllp{0.30\linewidth}lll@{}}
    \toprule
    \textbf{Rule id} & \textbf{Domain} & \textbf{Type} &
    \textbf{Condition (task types)} &
    \textbf{Effect} & \textbf{Prio} & \textbf{Min level} \\
    \midrule
    \multicolumn{7}{l}{\textit{Universal (both domains)}} \\[2pt]
    \texttt{exp\_autonomy\_cap}$^\star$
      & SW + MF & Cap
      & risk $\geq \tau_r$ \textbf{or} budget $< \tau_b$
      & cap $=$ \textsc{Supervised}  & dynamic & L1 \\
    \midrule
    \multicolumn{7}{l}{\textit{Software domain}} \\[2pt]
    \texttt{sw\_high\_judgment\_sup}
      & SW & Mode
      & Architecture Design, Code Review, Debugging
      & $\to$ \textsc{Supervised}    & 40 & L3 \\
    \texttt{sw\_requirements\_human}
      & SW & Hard
      & Requirements Analysis
      & force \textsc{Human}         & 15 & L4 \\
    \texttt{sw\_architecture\_human}
      & SW & Hard
      & Architecture Design
      & force \textsc{Human}         & 20 & L4 \\
    \texttt{sw\_testing\_sup}
      & SW & Mode
      & Testing, Refactoring
      & $\to$ \textsc{Supervised}    & 50 & L4 \\
    \midrule
    \multicolumn{7}{l}{\textit{Manufacturing domain}} \\[2pt]
    \texttt{mfg\_high\_impact\_sup}
      & MF & Mode
      & Quality Inspection, Pred.\ Maintenance,
        Process Programming, Process Optimisation
      & $\to$ \textsc{Supervised}    & 40 & L3 \\
    \texttt{mfg\_safety\_human}
      & MF & Hard
      & Safety Supervision
      & force \textsc{Human}         & 12 & L4 \\
    \texttt{mfg\_assembly\_sup}
      & MF & Mode
      & Assembly
      & $\to$ \textsc{Supervised}    & 45 & L4 \\
    \bottomrule
  \end{tabular}
  \end{adjustbox}
  \par\smallskip
  \begin{minipage}{\linewidth}
    \footnotesize\noindent
    Prio = rule priority (lower fires first); ``dynamic'' indicates that
    the rule is evaluated at runtime against per-subtask risk and budget
    signals rather than assigned a fixed rank.
    Hard rules override both the bandit and the autonomy cap by forcing
    a pure human-only or AI-only allocation.
    Mode rules constrain the bandit's action set by forcing a specific
    collaboration mode, but leave participation shares to the framework
    equations (\Cref{eq:copilot_share,eq:supervised_share}).
    $^\star$Cap thresholds $(\tau_r, \tau_b)$ vary by level; see
    \Cref{tab:gov_ladder_config}.
  \end{minipage}
\end{table}

This two-stage structure provides four desirable properties:
(\emph{i})~\emph{governance-first} --- the policy acts as a hard veto,
so no bandit optimisation can violate an organisational constraint;
(\emph{ii})~\emph{learning within bounds} --- the policy-defined autonomy cap
turns the autonomy spectrum into an explicit limit that the organisation
controls;
(\emph{iii})~\emph{continuous participation shares} --- the clipped equations
produce real-valued $\sigma$ that adapt to fatigue and complexity; and
(\emph{iv})~\emph{human state as an active signal} --- fatigue $f$ influences
the allocation decision at three points: it helps determine which mode is
selected, it affects how work is split between the human and the AI within
that mode, and it contributes to the reward received after execution. As a
result, the system can progressively move more load away from the human as
fatigue increases.

\subsection{From Allocation to Simulated Outcomes}
\label{sec:framework:execution}

\begin{figure}[htbp]
  \centering
  \begin{tikzpicture}[
      scale=0.89,
      transform shape,
      node distance=1.0cm and 0.9cm,
      loopbox/.style={rectangle, draw, rounded corners,
                      minimum width=2.7cm, minimum height=1.0cm,
                      align=center, font=\small},
      arr/.style={-{Stealth}, thick}
    ]
    \node[loopbox, fill=blue!10]   (input)  {1. Inputs\\subtask, state, policy};
    \node[loopbox, fill=orange!10, right=of input] (alloc) {2. Governed mode selection\\$\rightarrow (m,\sigma_{\mathrm{H}},\sigma_{\mathrm{AI}})$};
    \node[loopbox, fill=gray!10,   right=of alloc] (exec)  {3. Work execution\\mode semantics + shares};
    \node[loopbox, fill=green!10,  below=of alloc] (kpi)   {4. Outcomes and reward\\quality, time, cost, well-being};
    \node[loopbox, fill=red!10,    left=of kpi]    (state) {5. State update\\fatigue, trust, skill};

    \draw[arr] (input) -- (alloc);
    \draw[arr] (alloc) -- (exec);
    \draw[arr] (exec) -- (kpi);
    \draw[arr] (kpi) -- (state);
    \draw[arr] (state.north) |- (input.south);
  \end{tikzpicture}
  \caption{Five-step execution loop connecting governed allocation to the
    benchmark outcomes reported in \Cref{sec:results}.}
  \label{fig:execution_loop}
\end{figure}
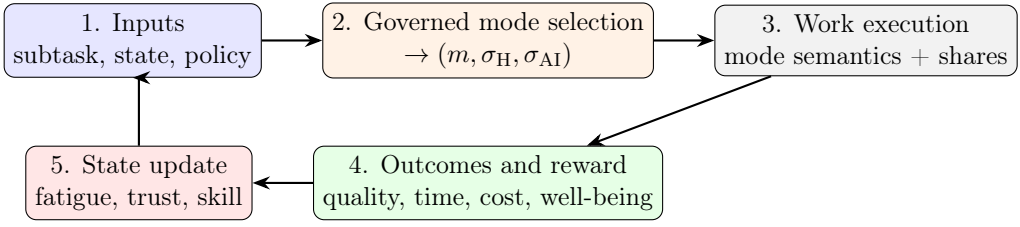

\Cref{fig:execution_loop} summarises the closed execution loop that connects
governed allocation to the benchmark outcomes.
Each cycle begins with a subtask $s$ and the current human state
$(f,\mathit{trust},\mathit{skill})$.
The allocation engine returns an allocation specification
$(m,\sigma_{\mathrm{H}},\sigma_{\mathrm{AI}})$; pure modes map to single-agent execution,
while shared modes distribute work according to the participation shares in
\Cref{sec:framework:modes}.
Execution then produces both operational outcomes (quality, time, and cost)
and human-impact consequences. The latter include the signals used to compute
composite well-being measures: fatigue burden, monotony, deskilling,
exclusion from cognitively valuable work, and any fatigue-relief benefit from
offloading. Together, these outcomes populate the benchmark KPIs
(\Cref{sec:setup:metrics}) and the scalar reward (\Cref{eq:reward}) that
updates the learner.
After each subtask, fatigue either accumulates or recovers, trust adjusts to
AI reliability, and skill decays when human participation remains too low
(\Cref{sec:framework:human}).
The benchmark therefore evaluates a full adaptive closed loop, not a static
assignment rule.

Simulated outcomes are generated by a \emph{parameterised stochastic model}
rather than a live agent.  For each subtask, quality is drawn from a Gaussian
whose mean is a weighted linear combination of five task-dimension scores
(repetitiveness, technical depth, creativity, ambiguity, and human-interaction
demand), with low noise ($\sigma=0.04$) for the AI agent and higher noise
($\sigma=0.07$) for the human agent to reflect individual variability.
Time taken is a deterministic fraction of the subtask baseline duration scaled
by task complexity, and cost follows directly from time and the agent's hourly
rate.  In shared modes the final quality combines both agents' contributions
with a synergy bonus proportional to their participation shares and the
subtask's technical depth, while time benefits from a partial-overlap factor.
This design makes the benchmark fully reproducible across seeds without
requiring live API calls.

\subsection{Human State Model}
\label{sec:framework:human}

A key distinguishing feature of \textsc{HAAS} relative to purely
technical allocation frameworks is its explicit representation of the
human agent's evolving state.
Allocation decisions that appear optimal at the beginning of a shift may
become harmful as fatigue accumulates, trust erodes, or sustained
disuse begins to degrade skill.
To model these dynamics, \textsc{HAAS} maintains three state variables for
each human agent that are updated after every subtask execution and feed
back into subsequent allocation cycles.

\paragraph{Fatigue}
Fatigue $f \in [0,1]$ accumulates at a base rate $\beta_f = 0.07$ per hour,
scaled by task complexity and a context-switching penalty
($\Delta f_{\mathrm{switch}} = 0.015$ per agent switch, capped at $0.08$
per cycle).
Recovery follows $f \leftarrow f - \beta_r \cdot t_{\mathrm{rest}}$
($\beta_r = 0.12$), with a chronic residual component ($\lambda = 0.18$)
that prevents full within-session recovery.
Parameters are heuristic defaults calibrated for qualitative consistency
with the cognitive fatigue literature~\citep{Warm2008,Yerkes1908,Endsley1995}.

\paragraph{Trust}
Trust $\mathit{trust} \in [0,1]$ decays on errors (rate $0.08$) and grows on
successes (rate $0.02$), with a floor of $0.35$ and damping of $0.60$.
The asymmetric rates reflect findings that trust repairs more slowly than it
degrades~\citep{Lee2004,Bucinca2021}.
Trust modulates the human's willingness to accept AI-led modes; low trust
increases the effective cost of \textsc{Supervised} and \textsc{Autonomous}
assignments.

\paragraph{Deskilling}
If the AI-execution fraction exceeds $\rho_{\mathrm{desk}} = 0.80$ for three
or more consecutive sprints, a deskilling rate of $0.003$ per cycle is applied,
operationalising the automation-complacency phenomena documented by
\citet{Parasuraman2010b} and \citet{Bainbridge1983}.
A tutor-mode intervention reactivates human tasks to prevent further loss.

Across all three state models, the parameter values are intended to capture
plausible dynamics grounded in prior literature rather than to provide
validated measurements of human behaviour (see \Cref{sec:discussion:limits});
key defaults are listed in \Cref{tab:params}.

To evaluate \textsc{HAAS} empirically, we instantiate the closed loop
described above (\Cref{fig:execution_loop}) in a reproducible simulation benchmark. The following section
details the simulation platform, the two application domains, the allocation
strategies compared, the outcome metrics, and the statistical protocol used
to assess robustness.

%% ================================================================
\section{Experimental Setup}
\label{sec:setup}
%% ================================================================

The following subsections describe the platform on which all experiments were
run (\Cref{sec:setup:platform}), the two application domains and their
scenarios (\Cref{sec:setup:domains}), the allocation strategies included in
the comparison (\Cref{sec:setup:strategies}), the outcome metrics
(\Cref{sec:setup:metrics}), and the statistical protocol used to assess
result robustness (\Cref{sec:setup:stats}).

\subsection{Simulation Platform}
\label{sec:setup:platform}

All experiments were conducted on the \textsc{Human--AI Symbiosis Studio},
a Python intelligent allocation system built with Streamlit (interactive
dashboard), Pandas/NumPy/SciPy (analytics), Plotly (visualisation), Pydantic v2
(immutable data models), and SQLite (persistent storage).
The platform is available for evaluation purposes
(see Software Availability Statement).

\begin{figure}[htbp]
  \centering
  \includegraphics[width=0.96\linewidth,height=0.68\textheight,keepaspectratio]{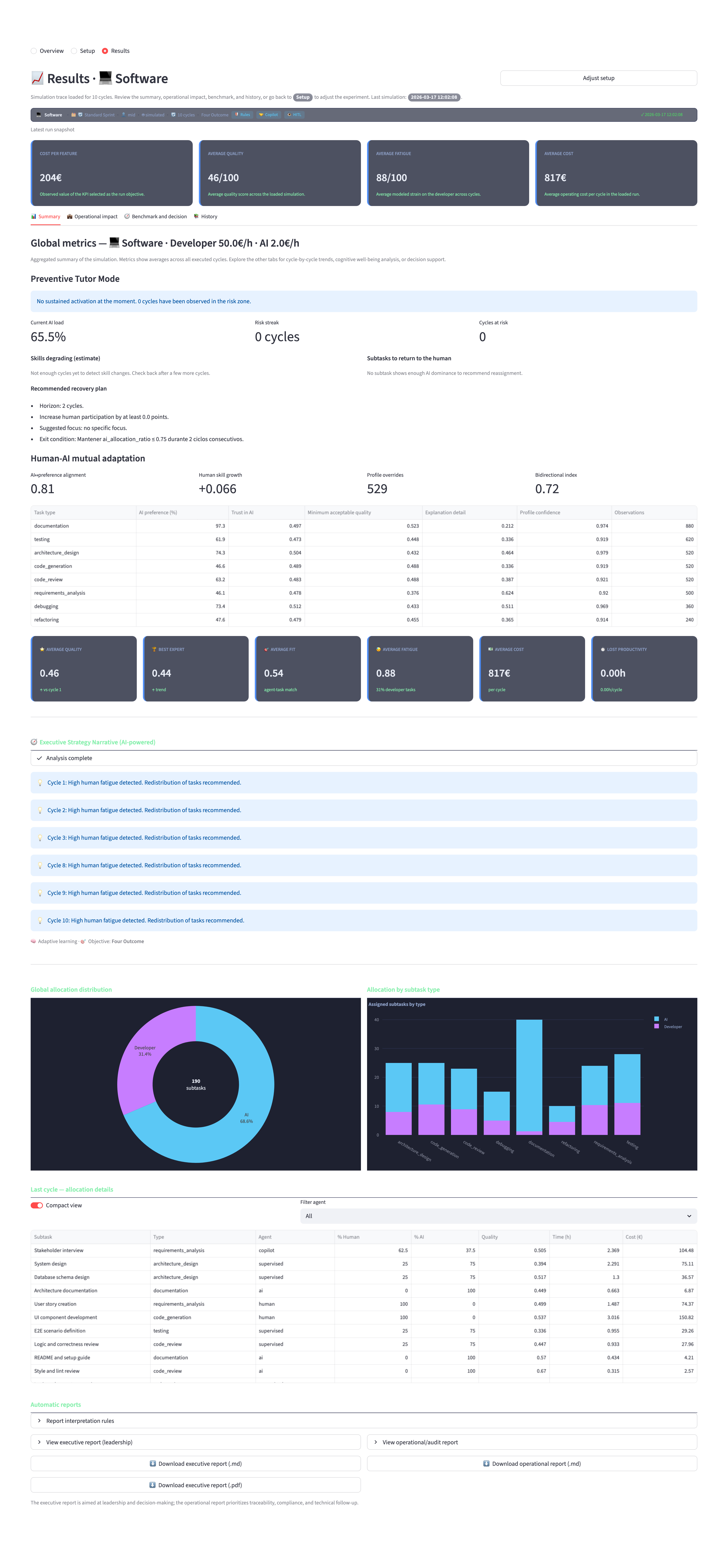}
  \caption{Screenshot of the \textsc{Human--AI Symbiosis Studio}
    dashboard (KPI summary view, manufacturing domain). The panel shows
    sprint-level KPIs, collaboration-mode distribution, and allocation history
    charts generated during a benchmark run.}
  \label{fig:dashboard}
\end{figure}

The Streamlit dashboard exposes five views: setup wizard, KPI summary,
allocation history, benchmark comparison, and a decision-support panel
rendering human state alongside governance-filtered recommendations.
\Cref{fig:dashboard} shows the KPI summary view for a manufacturing
simulation run, illustrating the sprint-level KPI table, collaboration-mode
distribution, and allocation-history charts.
Experiments are reproducible via fixed seeds and a command-line A/B runner;
the repository tuning workflow used to calibrate benchmark defaults is
summarised in \Cref{tab:tuning_appendix}.
Each benchmark run evaluates the framework in a single-human/single-AI
configuration, chosen to isolate allocation logic and governance effects
without altering the five collaboration modes defined by the framework.
Extending the benchmark to multi-actor coordination is left for future work.

Unless stated otherwise, the main benchmark uses a single
\emph{four-outcome} reward profile that balances quality, time, cost, and
well-being with weights 0.30, 0.20, 0.10, and 0.40, respectively.
The well-being component is itself a weighted composite of five sub-signals:
\begin{equation}
  \text{WB}(t) = 1 - \omega_f F(t) - \omega_m M(t) - \omega_d D(t) - \omega_e E(t) + \omega_r R(t),
  \label{eq:wellbeing}
\end{equation}
where the sub-weights are $\omega_f = 0.35$, $\omega_m = 0.20$, $\omega_d = 0.25$,
$\omega_e = 0.10$, $\omega_r = 0.10$.
The five sub-signals are defined as follows.
$F(t) \in [0,1]$ is the current normalised fatigue level (updated per the
dynamics in \Cref{sec:framework:human}).
$M(t) \in [0,1]$ is the monotony signal, defined as the fraction of the last
five subtasks assigned to the same agent without mode variation.
$D(t) \in [0,1]$ is the cumulative deskilling increment, activated when the
AI-execution fraction exceeds $\rho_{\mathrm{desk}} = 0.80$ for three or
more consecutive sprints (\Cref{sec:framework:human}).
$E(t) \in [0,1]$ is the cognitive exclusion signal, proportional to the
fraction of high-value subtasks (high technical depth or creativity) assigned
exclusively to AI in the current sprint.
$R(t) \in [0,1]$ is the shared-mode relief bonus, equal to the fraction of
subtasks executed in a shared mode (\textsc{Copilot}, \textsc{Peer}, or
\textsc{Supervised}) that reduced the human's predicted fatigue increment
relative to a pure-human assignment.
The oracle baseline is reported only as a hindsight upper bound on the scalar
reward objective in \Cref{eq:reward}, not on every KPI column individually: it
selects, for each subtask, the mode that would have maximised that reward if
the true outcome had already been known. Alternative reward priorities are
reported separately in \Cref{sec:results:sensitivity}.
Note that this oracle is \emph{mode-optimal}: it ranges over all five
collaboration modes and is therefore a strictly different reference from the
cumulative regret floor, which is defined in \Cref{sec:setup:metrics}
relative to the two pure counterfactual baselines (pure-human and pure-AI)
only. A mode-optimising oracle can itself accumulate positive regret under
the two-baseline definition, as the results tables confirm.

\subsection{Domains and Scenarios}
\label{sec:setup:domains}

To assess whether \textsc{HAAS} generalises across structurally different
work contexts, we instantiate the framework in two \emph{domains}
(software engineering and manufacturing) that differ in task granularity,
cost structure, and safety constraints.
Each domain contains four \emph{scenarios} (\Cref{tab:domains}), but the
main benchmark tables focus on the standard scenario per domain
(\emph{Standard Sprint} for software; \emph{Standard Production} for
manufacturing), using a common horizon of 8 cycles and 4 subtasks per cycle
to preserve cross-domain comparability.

\begin{table}[H]
  \centering
  \caption{Scenario catalog used by the \textsc{HAAS} benchmark battery.}
  \label{tab:domains}
  \footnotesize
  \begin{adjustbox}{max width=\linewidth}
  \begin{tabular}{ll>{\raggedright\arraybackslash}p{7.5cm}}
    \toprule
    \textbf{Domain} & \textbf{Scenario} & \textbf{Dominant pressure / emphasis} \\
    \midrule
    Software
      & Standard Sprint    & Balanced requirements, coding, review, and testing. \\
    Software
      & High Complexity    & Architecture-heavy work where human expertise is more decisive. \\
    Software
      & Maintenance        & Debugging, refactoring, review, and maintenance-oriented work. \\
    Software
      & Deadline Crunch    & Higher task pressure with faster human-fatigue accumulation. \\
    \midrule
    Manufacturing
      & Standard Production & Balanced operation, inspection, and maintenance. \\
    Manufacturing
      & Quality Crisis      & Defect analysis and process correction. \\
    Manufacturing
      & Scheduled Stop      & Maintenance and cobot reintegration. \\
    Manufacturing
      & New Product Ramp-Up & Ramp pressure with operator fatigue buffering. \\
    \bottomrule
  \end{tabular}
  \end{adjustbox}
\end{table}

The software domain uses a cost profile of €50/h for human agents and €2/h
for AI, reflecting current LLM API economics.
The manufacturing domain uses €32/h and €8/h respectively, reflecting
collaborative robot operating costs.
Human profiles range from junior/operator to senior/engineer, varying initial
skill and fatigue resistance.

\subsection{Allocation Strategies}
\label{sec:setup:strategies}

To make the performance of the three learned allocators easier to interpret,
the benchmark also includes simpler comparison strategies. These range from
basic heuristics to fixed human-only or AI-only assignments, as well as an
oracle upper bound.

The main benchmark reports the following strategy families:

\begin{enumerate}
  \item \textbf{Learned allocators}: UCB1, LinUCB, and Thompson Sampling.
    UCB1 and LinUCB are reported with policies enabled and disabled;
    Thompson Sampling is reported with policies enabled.
  \item \textbf{Heuristic baselines}: an affinity heuristic and a random
    allocator.
  \item \textbf{Managerial baseline}: a human+scheduler condition seeded by
    a manager-defined initial configuration, instantiated here as first-cycle
    affinity preferences.
  \item \textbf{Fixed baselines}: Fixed-human and AI-only.
  \item \textbf{Diagnostic comparator}: an oracle counterfactual baseline used
    as a reference point rather than as a feasible deployment policy.
\end{enumerate}

All bandit strategies begin with the same heuristic policy for the first
$T_{\mathrm{explore}} = 3$ sprints. This short warm-up phase gives each
strategy an initial set of observations before the learned bandit policy takes
over. The framework also supports a human-defined initial configuration via
the setup wizard, but that option is evaluated separately in the benchmark as
the managerial baseline rather than used to seed the learned allocators.

\subsection{Outcome Metrics}
\label{sec:setup:metrics}

We distinguish three families of outcome metrics. First, we report
\emph{operational KPIs} (key performance indicators): the main performance
measures used to judge how well each domain is operating. Because the two
domains model different kinds of work, these KPIs are domain-specific.
Second, we report \emph{cross-domain human-impact indicators}, which track
how allocation choices affect fatigue, trust, skill retention, and human
participation across both settings. Third, we report \emph{cumulative
allocation regret}, which serves as a learning diagnostic for the adaptive
strategies.

\paragraph{Operational KPIs}
In the software domain, the primary KPIs are lead time (hours/sprint), defect
escape rate (\%), rework percentage (\%), cost per feature (€), hybrid
subtask percentage, and fatigue avoided via copilot assistance. In the
manufacturing domain, the corresponding KPIs are Overall Equipment
Effectiveness (OEE) proxy (0--1), scrap rate
(\%), safety incident rate, cost per batch (€), stockout events per shift,
machine overload hours, hybrid subtask percentage, and fatigue avoided via
cobot assistance.

\paragraph{Cross-domain human-impact indicators}
To make the human consequences of allocation visible across both domains, we
also report end-of-sprint fatigue level ($f$), cumulative deskilling
increments $\sum \delta_d$, trust level, and human participation percentage.

\paragraph{Cumulative allocation regret}
As a learning diagnostic, the benchmark also reports cumulative allocation
regret. For each subtask, regret is the non-negative gap between the realized
reward and the better of the two pure counterfactual baselines (pure-human
and pure-AI execution under the same conditions). This floor at zero ensures
that when a shared collaboration mode outperforms both pure alternatives, the
contribution to regret is zero rather than negative.
The oracle strategy reported in the results tables
(\Cref{tab:software_results,tab:mfg_results}) is a separate, mode-optimal
hindsight comparator that selects among all five collaboration modes; it is
not the same reference as the two-baseline regret floor defined here, and
it is therefore possible for the oracle to accumulate positive regret under
this definition.

\paragraph{Governance screens}
To characterise \emph{how the policy layer changes the set of operationally
viable configurations}, the benchmark applies three nested diagnostic screens:
\begin{itemize}
  \item \emph{Acceptable}: five basic safeguards must hold simultaneously:
    mean quality remains above the minimum threshold, mean fatigue remains
    below the fatigue ceiling, estimated deskilling risk stays below its
    maximum bound, overall human participation remains above the minimum
    floor, and no governance violations occur.
  \item \emph{Reasonable}: all \emph{Acceptable} conditions hold, and two
    additional work-design conditions are satisfied: human work does not
    become excessively monotonous, and the allocation mix preserves a
    meaningful level of shared Human--AI collaboration while limiting fully
    autonomous execution on high-value tasks.
  \item \emph{Responsible}: all \emph{Reasonable} conditions hold, and
    capability-retention constraints are also satisfied: humans retain a
    minimum share of participation in high-value work, and risky tasks
    continue to involve shared Human--AI execution.
\end{itemize}
The screen definitions above specify the diagnostic conditions used in the
benchmark. Key algorithmic and human-state defaults are enumerated in the
simulation parameter reference (\Cref{tab:params}).

The purpose of these screens is diagnostic rather than prescriptive.
They do not constitute deployment recommendations or normative labels for
organisational practice.
Instead, they indicate whether a governance configuration keeps an
increasingly broad set of operational, human-impact, and capability-retention
objectives within bounds at the same time.
The contract-sensitivity analysis in \Cref{app:contract} then shows how the
set of screen-passing strategies changes when the reward profile is varied,
making explicit that the set of viable operating points depends on the
governance contract and decision objective being imposed.

\subsection{Statistical Protocol}
\label{sec:setup:stats}

The benchmark is designed primarily for comparative evaluation of allocation
patterns rather than for exhaustive hypothesis testing across all strategies.
Each strategy--scenario combination is run with 30 independent seeds and the
main tables report cross-seed means to make operating-point differences easy to
inspect.

To avoid a large multiple-comparison exercise on synthetic benchmark outputs,
only one pairwise comparison is tested inferentially: LinUCB with policies
disabled versus AI-only. This comparison was specified in advance because
AI-only defines the efficiency frontier in the audited standard scenarios,
whereas LinUCB\,+\,off is the strongest human-participatory configuration by
descriptive regret ranking.

The comparison is evaluated with a two-sided Wilcoxon signed-rank test using
the 30 paired seed outcomes, with $\alpha = 0.05$. Effect size is reported as
$r = Z/\sqrt{N}$, where $Z$ is the Wilcoxon normal-approximation statistic and
$N$ is the number of pairs. All remaining comparisons are interpreted
descriptively through cross-seed means, seed-level consistency, and the
sensitivity analyses reported later in the paper; seed-level dispersion for
the pre-specified AI-only versus LinUCB\,+\,off contrast is reported in
\Cref{tab:dispersion_appendix}.

The inferential result should be read as supportive evidence about simulation
stability under the specified benchmark settings, not as field-confirmatory
evidence about organisational deployment performance.

%% ================================================================
\section{Results}
\label{sec:results}
%% ================================================================

This section opens with the paper's main systems result: governance severity
can be benchmarked as a tunable design variable rather than treated as a
binary policies-on versus policies-off switch. The section is organised in
two blocks. It first develops the governance result in three steps: the
Governance Ladder benchmark in the audited standard scenarios, a
cross-scenario portability check of the preferred governance level, and a
long-horizon stability check. It then situates that result within the
broader benchmark by reporting strategy comparisons in the standard
scenario, portability across scenarios, shifts in collaboration modes,
human-state effects, component ablations, and contract sensitivity.

%% --- Governance Ladder subsections (formerly external \input) ---

\subsection{Governance Ladder: How Much Governance Is Needed?}
\label{sec:results:gov_ladder}

{\bfseries Context.}
The binary policies-on versus policies-off comparison is informative, but it
does not show how much governance is needed before the system moves from
useful authority control to over-restriction.
To make that trade-off explicit, we ran a five-level \emph{Governance Ladder}
benchmark with a fixed learner (\textsc{LinUCB}) and progressively stricter
\texttt{PolicyEngine} configurations.
L0 removes the policy layer entirely; L1 enables only a permissive autonomy
cap; L2 corresponds to the calibrated baseline governance setting; L3 adds
broader supervision rules for high-impact tasks; and L4 applies a
compliance-first configuration in which supervised or human-owned execution
becomes the dominant ceiling.
The two numeric thresholds are benchmark parameters on a 0--1 scale. The
\emph{risk threshold} is compared against the task-risk score computed from
subtask complexity and criticality; if the score exceeds that threshold, AI
autonomy is restricted. The \emph{autonomy-budget threshold} is compared
against the autonomy budget
($0.45\,\text{experience} + 0.25\,\text{process maturity} + 0.30\,(1 - \text{risk})$);
if the budget falls below that threshold, autonomy is also restricted. Lower
risk thresholds and higher budget thresholds therefore make the autonomy cap
easier to trigger and thus make governance more restrictive.

For this benchmark, \emph{Objective} denotes the domain-specific target KPI
used to rank governance levels under the active contract: cost per feature in
software and cost per batch in manufacturing. Both are minimization
objectives, so lower values are better.

{\bfseries Findings.}
\textit{Main result.}
Increasing governance intensity monotonically redistributes work away from
\textsc{Autonomous} execution and toward \textsc{Supervised} execution, but
the best governance level depends on the domain rather than being universal.
In software, the best 8-cycle operating point remains L0, whereas in
manufacturing the best standard-scenario operating point appears at L3.

{\bfseries Evidence.}
\Cref{tab:gov_ladder_insert_main,tab:gov_ladder_insert_interventions} report
the audited standard scenarios across 30 seeds.
In software, \textsc{Autonomous} execution falls from 55.5\% at L0 to
22.7\% at L2 and to 0.0\% at L3--L4, while \textsc{Supervised} execution
rises from 16.6\% at L0 to 54.4\% at L2 and 87.7\% at L3.
The key authority shift therefore occurs between L1 and L2.
Operationally, however, L0 remains best on the benchmark objective
(161.48), and the largest additional penalty appears at the top of the
ladder: objective value worsens from 199.05 at L3 to 243.77 at L4, while
fatigue rises from 0.757 to 0.956.

Manufacturing shows a different pattern.
\textsc{Autonomous} execution falls from 40.1\% at L0 to 14.7\% at L2 and
3.1\% at L3--L4, while \textsc{Supervised} execution rises from 18.2\% at
L0 to 52.9\% at L2 and 83.2\% at L3.
Unlike software, the move to stronger governance is not purely a cost:
objective value improves from 147.13 at L0 to 139.07 at L3, and fatigue
falls from 0.926 to 0.842.
L4 further increases rule-forced decisions but does not improve the
objective over L3.

\begin{table}[!ht]
  \centering
  \caption{Governance Ladder results in the audited standard scenarios (30 seeds, 8 cycles, \textsc{LinUCB}). Objective denotes the domain-specific target KPI: cost per feature in software and cost per batch in manufacturing; lower values are better.}
  \label{tab:gov_ladder_insert_main}
  \footnotesize
  \setlength{\tabcolsep}{3pt}
  \renewcommand{\arraystretch}{1.03}
  \begin{adjustbox}{max width=\linewidth}
  \begin{tabular}{llrrrrr}
    \toprule
    \textbf{Domain} & \textbf{Level} & \textbf{Objective} & \textbf{Lead time} & \textbf{Quality} & \textbf{Fatigue} & \textbf{Cum.\ regret} \\
    \midrule
    Software & L0 & 161.48 & 23.13 & 0.502 & 0.734 & 10.68 \\
    Software & L1 & 162.47 & 23.20 & 0.503 & 0.737 & 10.69 \\
    Software & L2 & 184.86 & 24.00 & 0.458 & 0.743 & 12.96 \\
    Software & L3 & 199.05 & 24.33 & 0.438 & 0.757 & 14.19 \\
    Software & L4 & 243.77 & 26.93 & 0.414 & 0.956 & 17.49 \\
    \midrule
    Manufacturing & L0 & 147.13 & 24.76 & 0.463 & 0.926 & 10.94 \\
    Manufacturing & L1 & 144.62 & 24.51 & 0.465 & 0.909 & 10.64 \\
    Manufacturing & L2 & 151.31 & 24.96 & 0.452 & 0.889 & 10.98 \\
    Manufacturing & L3 & 139.07 & 23.58 & 0.460 & 0.842 &  9.56 \\
    Manufacturing & L4 & 144.65 & 24.16 & 0.460 & 0.886 &  9.82 \\
    \bottomrule
  \end{tabular}
  \end{adjustbox}
\end{table}

\begin{table}[!ht]
  \centering
  \caption{Governance intervention intensity and mode redistribution across the ladder (30 seeds, 8 cycles, \textsc{LinUCB}).}
  \label{tab:gov_ladder_insert_interventions}
  \footnotesize
  \setlength{\tabcolsep}{2.5pt}
  \begin{adjustbox}{max width=\linewidth}
  \begin{tabular}{llrrrrr}
    \toprule
    \textbf{Domain} & \textbf{Level} & \textbf{Policy caps} & \textbf{Policy forced} & \textbf{Sup. share} & \textbf{Auto. share} & \textbf{Human part.} \\
    \midrule
    Software & L0 & 0.00\% & 0.00\% & 16.56\% & 55.50\% & 22.96\% \\
    Software & L1 & 0.59\% & 0.00\% & 17.12\% & 55.04\% & 23.01\% \\
    Software & L2 & 38.87\% & 0.00\% & 54.44\% & 22.73\% & 29.11\% \\
    Software & L3 & 50.00\% & 34.42\% & 87.66\% &  0.00\% & 34.17\% \\
    Software & L4 & 28.57\% & 65.58\% & 70.13\% &  0.00\% & 40.57\% \\
    \midrule
    Manufacturing & L0 &  0.00\% &  0.00\% & 18.24\% & 40.13\% & 33.50\% \\
    Manufacturing & L1 &  4.50\% &  7.63\% & 21.91\% & 38.70\% & 32.58\% \\
    Manufacturing & L2 & 42.08\% &  7.63\% & 52.90\% & 14.73\% & 36.10\% \\
    Manufacturing & L3 & 17.56\% & 70.23\% & 83.21\% &  3.05\% & 34.03\% \\
    Manufacturing & L4 &  7.63\% & 86.26\% & 80.92\% &  3.05\% & 35.21\% \\
    \bottomrule
  \end{tabular}
  \end{adjustbox}
\end{table}

\begin{figure}[htbp]
  \centering
  \includegraphics[width=0.92\linewidth,height=0.36\textheight,keepaspectratio]{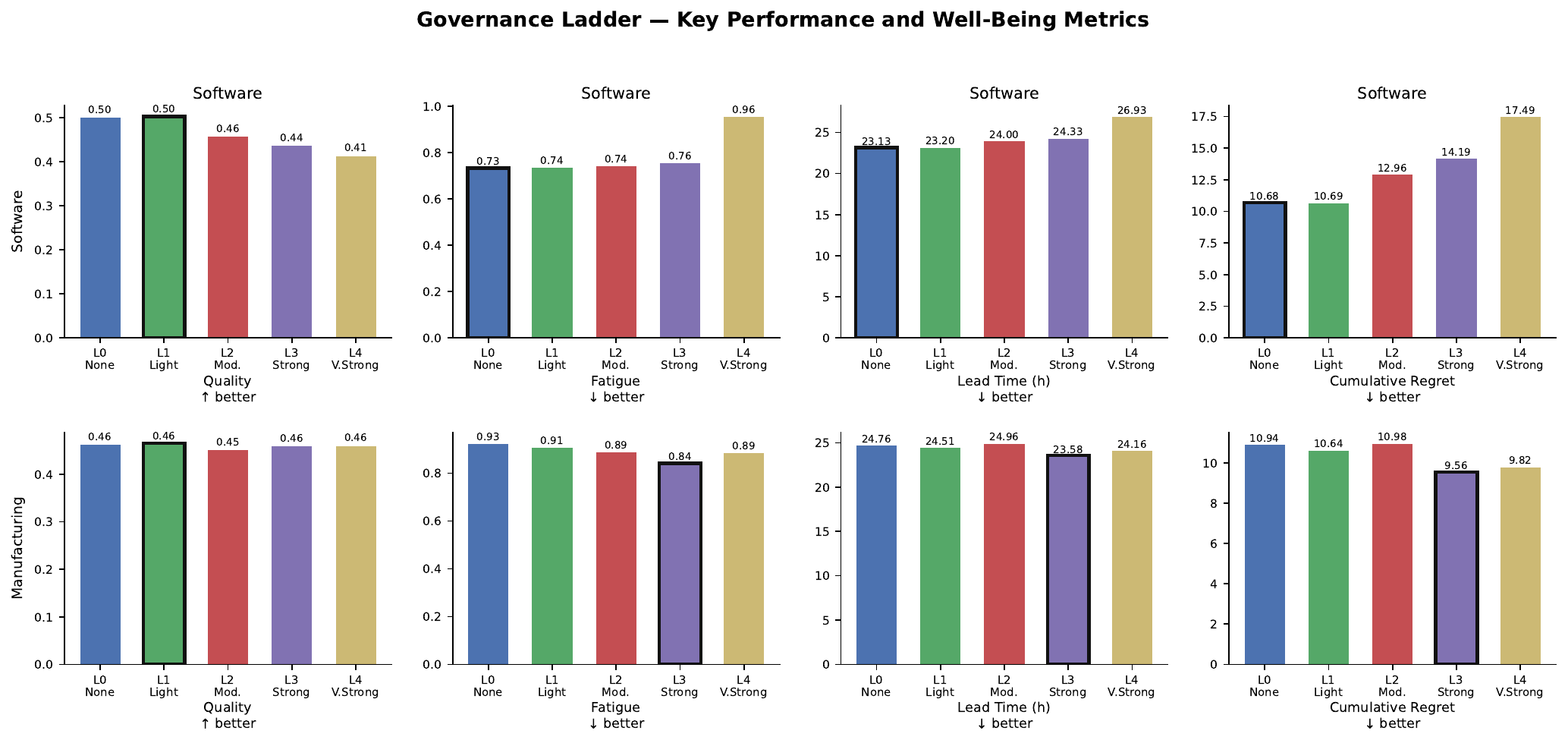}
  \caption{Governance Ladder --- quality, fatigue, lead time, and cumulative
             regret per level and domain (30 seeds, 8 cycles, \textsc{LinUCB}).
             Thicker bar border marks the domain optimum.
             Manufacturing optimal: L3. Software optimal: L0.}
  \label{fig:gov_ladder_main_metrics}
\end{figure}

\begin{figure}[htbp]
  \centering
  \includegraphics[width=0.92\linewidth,height=0.30\textheight,keepaspectratio]{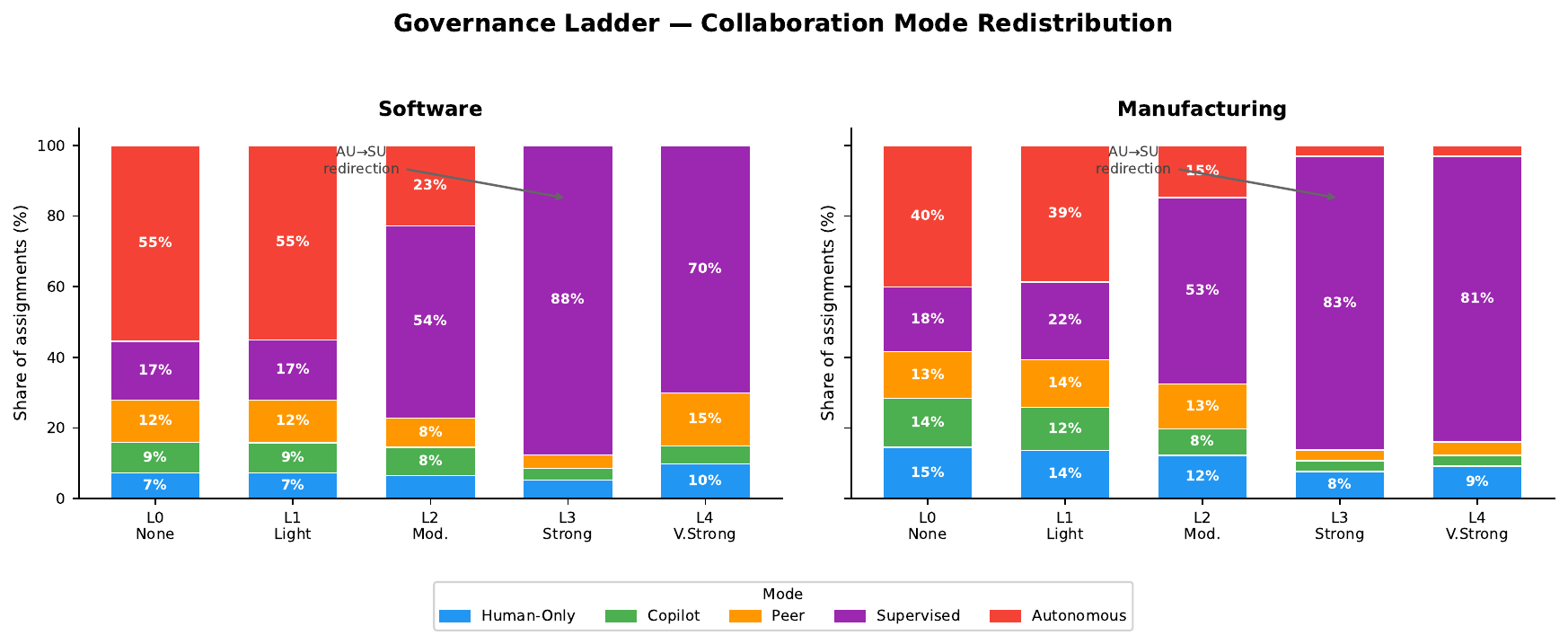}
  \caption{Collaboration mode redistribution across governance levels.
             \textsc{Autonomous} execution (red) collapses monotonically as governance
             tightens; \textsc{Supervised} (purple) absorbs the difference.
             \textsc{Human-Only} does \emph{not} grow, confirming that governance
             converts AI autonomy into supervised collaboration, not human
             replacement.}
  \label{fig:gov_ladder_modes}
\end{figure}

{\bfseries Interpretation.}
The practical value of governance lies in the middle region of the ladder.
In software, L2 is the last level that materially redistributes authority
without collapsing the benchmark into a compliance-first regime.
In manufacturing, L3 is the clearest ``useful control'' point: it achieves
the strongest authority retention together with the best objective value in
the audited standard scenario (\Cref{fig:gov_ladder_main_metrics}).
The mode redistribution (\Cref{fig:gov_ladder_modes}) shows that this
improvement is structural: governance systematically redirects autonomous
AI assignments into supervised collaborations rather than simply restoring
human-only execution.
Taken together, these results sharpen the binary policies-on comparison: the
useful level of governance is not fixed but depends on the domain, and the
ladder provides a systematic way to locate it before deployment.

\subsection{Scenario-Dependent Governance Level}
\label{sec:results:gov_ladder_portability}

{\bfseries Context.}
The standard scenarios identify whether a graded governance regime is useful
in the two audited cases, but they do not show whether the same governance
level transfers across contexts.
To test that point, we repeated the ladder over all eight repository
scenarios using 10 seeds per scenario.

{\bfseries Findings.}
\textit{Main result.}
The best governance level is scenario-dependent.
No single ladder level dominates the full scenario battery.

{\bfseries Evidence.}
\Cref{tab:gov_ladder_insert_winners} summarises the winner by scenario.
L0 is best in four scenarios: software \emph{Standard Sprint},
\emph{Maintenance}, and \emph{Deadline Crunch}, plus manufacturing
\emph{New Product Ramp-Up}.
L3 is best in the other four: software \emph{High Complexity}, and
manufacturing \emph{Quality Crisis}, \emph{Standard Production}, and
\emph{Scheduled Stop}.
The resulting split is exactly 4:4 between the no-governance and
strong-governance ends of the ladder.
Because this portability sweep uses a separate 10-seed design, the objective
values in \Cref{tab:gov_ladder_insert_winners} are not expected to match
exactly the 30-seed audited standard-scenario means reported earlier in
\Cref{tab:gov_ladder_insert_main}; here the inferential target is the winning
governance level within each scenario.

\begin{table}[H]
  \centering
  \caption{Best Governance Ladder level by scenario (10 seeds, 8 cycles, \textsc{LinUCB}). Objective values are from the portability sweep and therefore are not directly comparable to the 30-seed audited means in \Cref{tab:gov_ladder_insert_main}.}
  \label{tab:gov_ladder_insert_winners}
  \footnotesize
  \setlength{\tabcolsep}{4pt}
  \begin{adjustbox}{max width=\linewidth}
  \begin{tabular}{lp{0.28\linewidth}lrrr}
    \toprule
    \textbf{Domain} & \textbf{Scenario} & \textbf{Best level} & \textbf{Objective} & \textbf{Supervised share} & \textbf{Autonomous share} \\
    \midrule
    Software & Standard Sprint    & L0 & 162.27 & 16.82\% & 55.19\% \\
    Software & High Complexity    & L3 & 176.72 & 50.33\% & 32.87\% \\
    Software & Maintenance        & L0 & 133.92 & 14.10\% & 58.12\% \\
    Software & Deadline Crunch    & L0 & 177.47 & 17.98\% & 55.39\% \\
    Manufacturing & Standard Production & L3 & 139.06 & 83.21\% &  3.05\% \\
    Manufacturing & Quality Crisis      & L3 & 144.63 & 81.56\% &  5.16\% \\
    Manufacturing & Scheduled Stop      & L3 & 100.46 & 64.38\% & 21.04\% \\
    Manufacturing & New Product Ramp-Up & L0 & 145.35 & 17.83\% & 47.65\% \\
    \bottomrule
  \end{tabular}
  \end{adjustbox}
\end{table}

\begin{figure}[htbp]
  \centering
  \includegraphics[width=\linewidth,height=0.55\textheight,keepaspectratio]{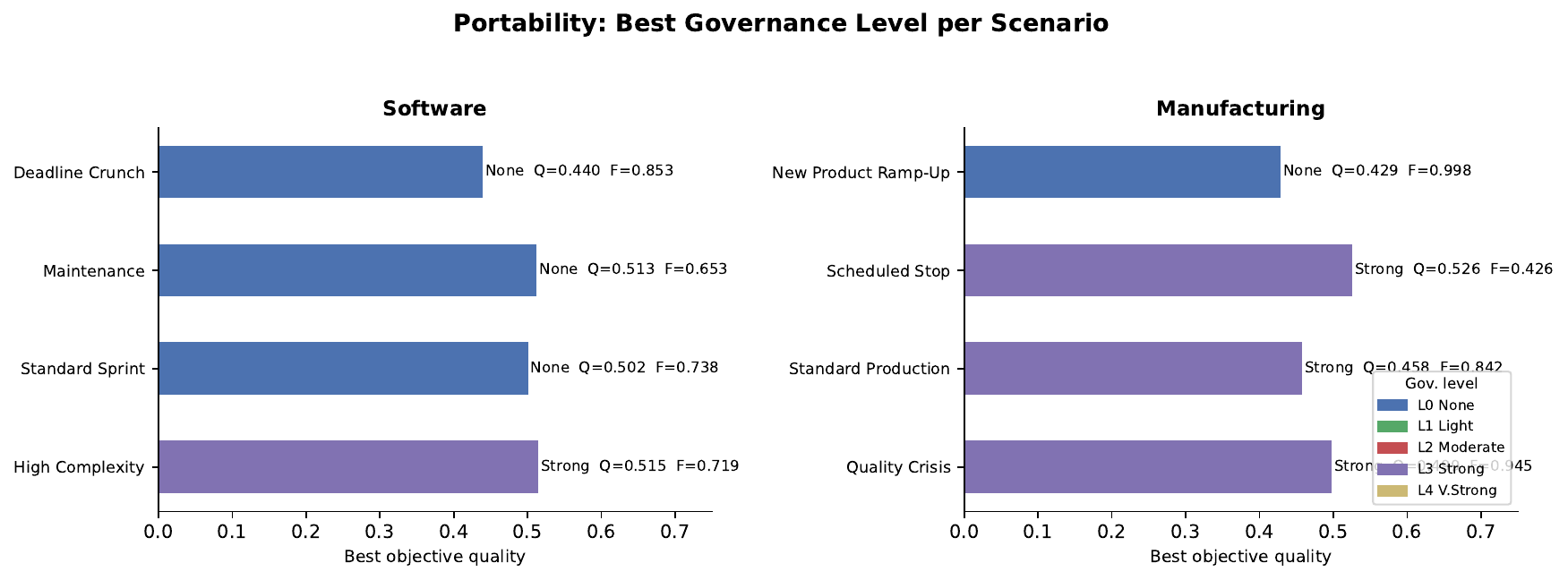}
  \caption{Best governance level per scenario (portability battery,
             10 seeds, 8 cycles). Bar length encodes the objective value at
             the winning level; colour encodes the level (L0 blue, L3 red).
             L0 and L3 each win exactly four scenarios.}
  \label{fig:gov_ladder_portability}
\end{figure}

{\bfseries Interpretation.}
\Cref{fig:gov_ladder_portability} sharpens the main claim of the paper:
the 4:4 split between L0 and L3 winners makes clear that it is not enough
to say that governance matters; one must also ask \emph{how much} governance
a given scenario can absorb before the control cost outweighs the authority
benefit.
The answer clearly depends on the scenario.

\subsection{Long-Horizon Governance Check}
\label{sec:results:gov_ladder_long}

{\bfseries Context.}
The main benchmark uses an 8-cycle horizon, which is sufficient to reveal
directional effects but still relatively short for a learning-based
allocator.
To test whether the ladder shape changes when the learner has more time to
adapt, we repeated the standard scenarios at 16 cycles for L0, L2, and L4.

{\bfseries Findings.}
\textit{Main result.}
Moderate governance becomes more competitive over longer horizons.
At 16 cycles, L2 improves over L0 in both audited standard scenarios,
whereas L4 remains over-restrictive.

{\bfseries Evidence.}
\Cref{tab:gov_ladder_insert_long} reports the long-horizon comparison.
In software \emph{Standard Sprint}, objective value falls from 159.45 at L0
to 157.08 at L2 while \textsc{Supervised} execution rises from 19.6\% to
52.3\%.
In manufacturing \emph{Standard Production}, objective value falls from
135.60 at L0 to 130.06 at L2, and fatigue declines from 0.857 to 0.733.
L4 remains substantially more restrictive in both domains, with no
corresponding gain over L2.

\begin{table}[H]
  \centering
  \caption{Long-horizon check in the audited standard scenarios (30 seeds, 16 cycles, \textsc{LinUCB}). Objective denotes the domain-specific target KPI: cost per feature in software and cost per batch in manufacturing; lower values are better.}
  \label{tab:gov_ladder_insert_long}
  \footnotesize
  \setlength{\tabcolsep}{4pt}
  \begin{adjustbox}{max width=\linewidth}
  \begin{tabular}{llrrrrr}
    \toprule
    \textbf{Domain} & \textbf{Level} & \textbf{Objective} & \textbf{Lead time} & \textbf{Quality} & \textbf{Fatigue} & \textbf{Supervised share} \\
    \midrule
    Software & L0 & 159.45 & 22.57 & 0.493 & 0.709 & 19.56\% \\
    Software & L2 & 157.08 & 22.18 & 0.473 & 0.610 & 52.26\% \\
    Software & L4 & 230.10 & 25.83 & 0.428 & 0.929 & 73.36\% \\
    \midrule
    Manufacturing & L0 & 135.60 & 23.79 & 0.472 & 0.857 & 21.09\% \\
    Manufacturing & L2 & 130.06 & 23.09 & 0.474 & 0.733 & 56.12\% \\
    Manufacturing & L4 & 139.23 & 23.77 & 0.472 & 0.831 & 83.33\% \\
    \bottomrule
  \end{tabular}
  \end{adjustbox}
\end{table}

\begin{figure}[htbp]
  \centering
  \includegraphics[width=0.96\linewidth,height=0.38\textheight,keepaspectratio]{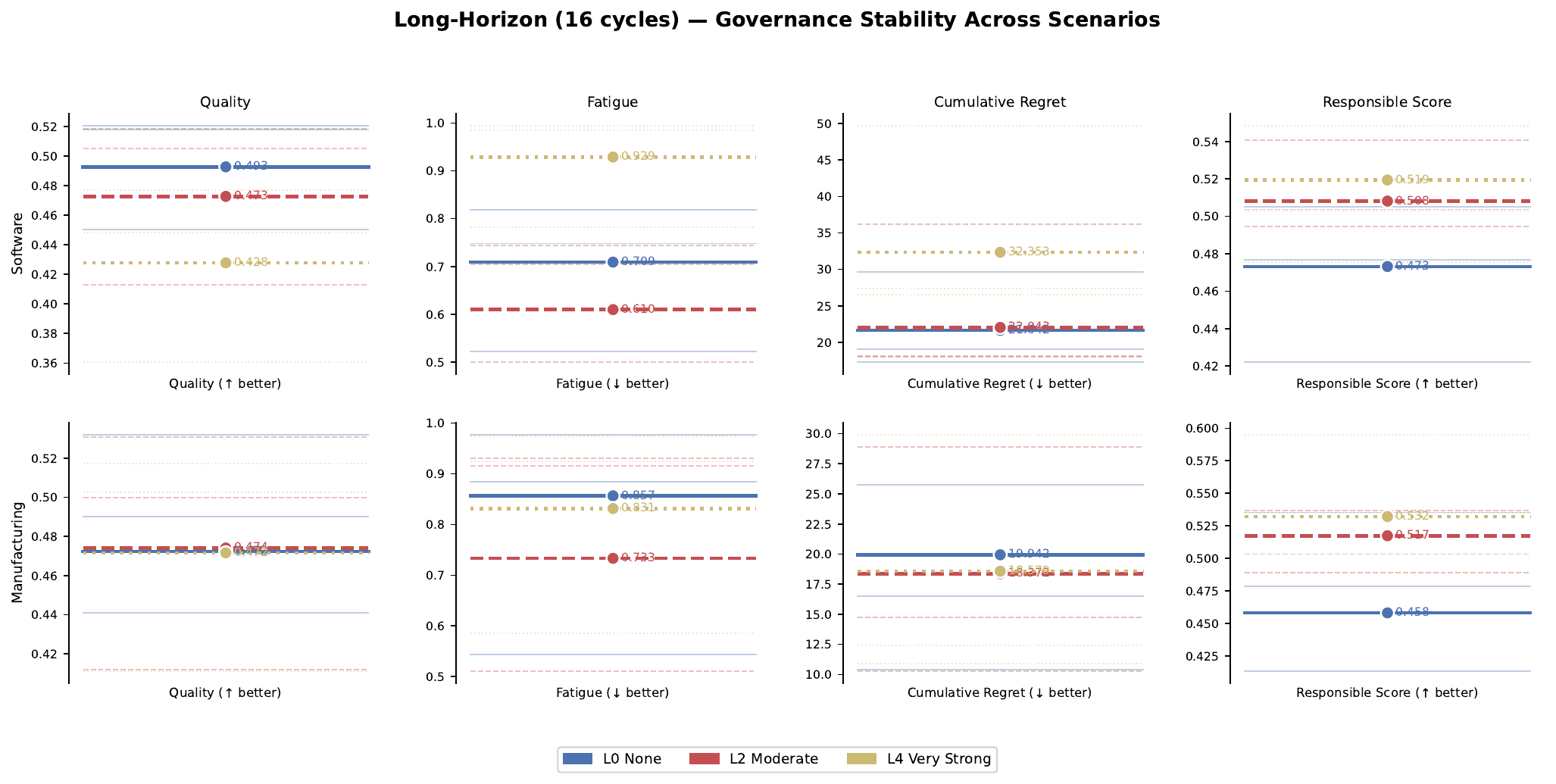}
  \caption{Long-horizon stability (16 cycles, 30 seeds) for L0, L2, and
             L4 on four outcome dimensions.
             Thick markers: standard scenario; faint lines: other scenarios
             in the same domain (context variation).
             Manufacturing: L2 improves over L0 on all axes; L4 remains
             over-restrictive despite high supervision share.
             Software: L4 compounds regret and fatigue; L2 achieves the
             most balanced profile.}
  \label{fig:gov_ladder_long}
\end{figure}

{\bfseries Interpretation.}
\Cref{fig:gov_ladder_long} shows that the governance effects are
\emph{persistent}, not transient.
The manufacturing advantage of strong governance strengthens over time;
the software penalty of over-governance compounds.
Some of the short-run cost of moderate governance is offset when the learner
has more cycles to acquire subtask-level preferences inside the governed action
space.
The ladder should therefore be read not only as a policy-severity analysis,
but also as a horizon-sensitive benchmark of governed learning.

\subsection{Strategy Comparison — Software Domain}
\label{sec:results:software}

{\bfseries Context.}
At the full-strategy benchmark level, the software domain has high task
heterogeneity,
the cost differential between human and AI agents is large (€50/h versus
€2/h), and no hard safety constraints force specific assignments, giving
learned allocators the widest discretion.

{\bfseries Findings.}
\textit{Main result.} In the software benchmark, AI-only is the efficiency
frontier, whereas LinUCB\,+\,off is the strongest strategy
that still preserves meaningful human participation.
(Throughout this section, \emph{AI-only} denotes the fixed-baseline strategy
that assigns every subtask to the AI agent alone, whereas
\textsc{Autonomous} denotes the collaboration mode within the five-mode
spectrum where $\sigma_{\mathrm{AI}} = 1.0$.
An AI-only baseline operates exclusively in \textsc{Autonomous} mode, but
the \textsc{Autonomous} mode can also be selected individually by governed
allocators for specific subtask types.)

{\bfseries Evidence.}
\Cref{tab:software_results} reports mean KPI values across 30 seeds for the
Standard Sprint scenario (8 cycles per run, post-calibration setting).

AI-only achieves the shortest lead time (16.63\,h/cycle), the lowest fatigue
(0.124), and the lowest regret (4.10). Under the active reward contract,
\textsc{HAAS} therefore does not surpass fixed AI-only execution on aggregate
throughput.

Among the human-participatory strategies, LinUCB\,+\,off is
the most favourable operating point by descriptive regret ranking. It stays close to the AI-only quality
level (0.502 vs.\ 0.505) while improving substantially on time, fatigue, and
regret relative to the other learned and heuristic alternatives
(pre-specified Wilcoxon: $W{=}0$, $p{<}0.001$, $r{=}0.87$, indicating that
AI-only achieved lower regret in every one of the 30 matched seed pairs).
LinUCB\,+\,on is slightly slower (24.00\,h) and lower in
quality (0.458), but it remains the strongest governance-enabled
configuration.

{\bfseries Interpretation.}
Two additional observations help interpret the table. First, the oracle
counterfactual (quality 0.528, fatigue 0.232) is a \emph{mode-optimal}
hindsight comparator: it selects the best of all five collaboration modes per
subtask with perfect foresight, and thereby shows that the simulator contains
higher quality states than most online strategies currently reach.
Importantly, the oracle is \emph{not} the reference used to compute
cumulative regret: regret is measured against the better of pure-human and
pure-AI only (see \Cref{sec:setup:metrics}).
In other words, the oracle is mode-optimal, not regret-minimal relative to
the AI-only baseline.
A mode-optimising oracle can therefore itself accrue positive regret under
that two-baseline definition (oracle regret = 4.71 vs.\ AI-only regret = 4.10
in this table), which explains why AI-only, a fixed pure strategy, can
outrank the oracle on the regret column.
Second, with only 8~sprints and a 3-sprint warm-up, bandits have limited time
to learn; the competitive performance of the affinity heuristic is therefore
consistent with a short-horizon benchmark
(\Cref{sec:discussion:future}).
\Cref{tab:software_results,tab:mfg_results} report cross-seed
means, sorted by cumulative regret; best values per column are shown in
\textbf{bold} and worst values in \textit{italic}.

\begin{table}[H]
  \centering
  \caption{Strategy comparison in the software Standard Sprint scenario.}
  \label{tab:software_results}
  \footnotesize
  \setlength{\tabcolsep}{4pt}
  \begin{tabularx}{\linewidth}{@{}>{\raggedright\arraybackslash}Xrrrrr@{}}
    \toprule
    \textbf{Strategy}
      & \textbf{\shortstack{Lead time\\(h/sprint)}}
      & \textbf{Quality}
      & \textbf{\shortstack{Cost\\(€)}}
      & \textbf{Fatigue}
      & \textbf{\shortstack{Cum.\\regret}} \\
    \midrule
    AI-only baseline                &  \textbf{16.63} & 0.505 & \textbf{154.57} & \textbf{0.124} &  \textbf{4.10} \\
    Oracle counterfactual baseline  &  18.36 & \textbf{0.528} & 273.54 & 0.232 &  4.71 \\
    \midrule
    LinUCB\,+\,off         &  23.13 & 0.502 & 645.91 & 0.734 & 10.69 \\
    LinUCB\,+\,on          &  24.00 & 0.458 & 739.44 & 0.743 & 12.96 \\
    UCB1\,+\,off           &  24.50 & 0.496 & 746.46 & 0.870 & 12.76 \\
    Affinity heuristic baseline    &  26.29 & 0.501 & 855.10 & 0.959 & 13.30 \\
    Thompson\,+\,on        &  26.10 & 0.444 & 895.45 & 0.927 & 16.10 \\
    UCB1\,+\,on            &  25.48 & 0.436 & 864.87 & 0.919 & 16.56 \\
    Human+scheduler baseline       &  25.73 & 0.449 & 759.19 & 0.822 & 16.57 \\
    \midrule
    Random baseline                &  29.57 & 0.356 & 1135.27 & 0.998 & 27.54 \\
    Fixed-human baseline           & \textit{42.12} & \textit{0.166} & \textit{2106.23} & \textit{1.000} & \textit{51.92} \\
    \bottomrule
  \end{tabularx}
  \par\smallskip
  \begin{minipage}{\linewidth}
    \footnotesize\noindent
    $n=30$ seeds (primes 11--151); values are cross-seed means.
    Wilcoxon signed-rank test (LinUCB\,+\,off vs AI-only, pre-specified):
    $W=0$, $p < 0.001$, $r = 0.87$ (large effect; AI-only achieved lower
    regret in all 30 matched seed pairs). The Oracle counterfactual is a
    \emph{mode-optimal} hindsight baseline (best of all five modes per subtask
    under perfect foresight); it achieves higher quality than AI-only
    (0.528 vs.\ 0.505) but accumulates higher regret (4.71 vs.\ 4.10).
    The latter occurs because regret is measured relative to the pure-AI floor:
    the oracle's occasional non-AI assignments incur positive regret under
    that reference even when they improve quality. The oracle is therefore
    \emph{not} a global performance ceiling on the regret column, and it is
    \emph{not} the reference for cumulative regret
    (see \Cref{sec:setup:metrics}).
  \end{minipage}
\end{table}

\subsection{Strategy Comparison — Manufacturing Domain}
\label{sec:results:manufacturing}

{\bfseries Context.}
The manufacturing domain differs structurally from software: it introduces
hard policy rules (human-only safety escalation, AI-only machine operations),
a narrower cost differential (€32/h vs.\ €8/h), and safety/OEE considerations
in the reward, all of which constrain the learner's degrees of freedom.

{\bfseries Findings.}
\textit{Main result.} The directional pattern is the same as software ---
AI-only at the efficiency frontier --- but governance effects are smaller and
the affinity heuristic is a more credible competitor.

{\bfseries Evidence.}
\Cref{tab:mfg_results} reports results for the Standard Production scenario.
AI-only gives the lowest lead time, fatigue, and regret (17.38\,h, 0.358,
3.48; Wilcoxon $W{=}0$, $p{<}0.001$, $r{=}0.87$ for LinUCB\,+\,off vs.\
AI-only, again unanimous across all 30 seed pairs).
The $W{=}0$ result matches the software domain exactly, reflecting that
AI-only achieved a strictly lower regret in every individual seed run in
both domains; given the large cost differential between agents, the
AI-only efficiency advantage is deterministic across seeds rather than
stochastic.
LinUCB\,+\,off is the strongest learned strategy by descriptive
regret ranking
(24.76\,h, 0.463 quality, 10.94 regret).

Crucially, the affinity heuristic performs differently here than in software:
it achieves competitive quality (0.493) and regret (10.22) but at the cost
of near-saturated fatigue (0.989).
This reflects the manufacturing subtask catalogue's higher density of physically
demanding, human-affinity tasks (assembly, precision inspection, floor
supervision) that the heuristic assigns to the human agent regardless of
fatigue state --- a pattern LinUCB avoids by adapting to the human-state
feedback signal.

\begin{table}[H]
  \centering
  \caption{Strategy comparison in the manufacturing Standard Production scenario.}
  \label{tab:mfg_results}
  \footnotesize
  \setlength{\tabcolsep}{4pt}
  \begin{tabularx}{\linewidth}{@{}>{\raggedright\arraybackslash}Xrrrrr@{}}
    \toprule
    \textbf{Strategy}
      & \textbf{\shortstack{Lead time\\(h/sprint)}}
      & \textbf{Quality}
      & \textbf{\shortstack{Cost\\(€)}}
      & \textbf{Fatigue}
      & \textbf{\shortstack{Cum.\\regret}} \\
    \midrule
    AI-only baseline                &  \textbf{17.38} & 0.496 & \textbf{259.54} & \textbf{0.358} &  \textbf{3.48} \\
    Oracle counterfactual baseline  &  20.37 & \textbf{0.528} & 387.38 & 0.563 &  4.54 \\
    \midrule
    Affinity heuristic baseline    &  26.90 & 0.493 & 659.74 & 0.989 & 10.22 \\
    LinUCB\,+\,off         &  24.76 & 0.463 & 588.54 & 0.926 & 10.94 \\
    LinUCB\,+\,on          &  24.96 & 0.452 & 605.24 & 0.889 & 10.98 \\
    UCB1\,+\,on            &  25.71 & 0.445 & 639.37 & 0.934 & 11.90 \\
    Thompson\,+\,on        &  26.97 & 0.441 & 689.12 & 0.988 & 13.02 \\
    UCB1\,+\,off           &  26.68 & 0.449 & 671.86 & 0.985 & 13.05 \\
    Human+scheduler baseline       &  28.85 & 0.408 & 733.76 & 0.998 & 16.39 \\
    \midrule
    Random baseline                &  29.13 & 0.372 & 771.86 & 0.999 & 19.82 \\
    Fixed-human baseline           & \textit{40.23} & \textit{0.230} & \textit{1264.62} & \textit{1.000} & \textit{36.03} \\
    \bottomrule
  \end{tabularx}
  \par\smallskip
  \begin{minipage}{\linewidth}
    \footnotesize\noindent
    $n=30$ seeds (primes 11--151); values are cross-seed means.
    Wilcoxon signed-rank test (LinUCB\,+\,off vs AI-only, pre-specified):
    $W=0$, $p < 0.001$, $r = 0.87$ (large effect).
  \end{minipage}
\end{table}

{\bfseries Interpretation.}
The practical interpretation of the on/off policy comparison in
\Cref{tab:mfg_results} is that governance does not improve throughput in this
benchmark slice. Policies-on is slightly slower and lower in quality than
LinUCB\,+\,off. In manufacturing, the binary policies-on result
should therefore be read as an explicit operating constraint rather than as a
free performance gain; the graded Governance Ladder analysis
(\Cref{sec:results:gov_ladder}) then shows that stronger governance levels can
still become beneficial under a different calibration of rule intensity.

\subsection{Cross-Scenario Portability}
\label{sec:results:portability}

{\bfseries Context.}
The audited standard-scenario tables above establish the baseline ranking per
domain. To check whether those rankings generalise, we swept all eight
scenarios with four focal conditions
(AI-only, LinUCB\,+\,on/off, affinity heuristic), 10 seeds per scenario.

{\bfseries Findings.}
\textit{Main result.} The AI-only efficiency frontier is stable; the best
human-participatory strategy is not.

{\bfseries Evidence.}
AI-only is the best objective-value strategy in all eight scenarios
shown in \Cref{tab:portability_main}. Among human-participatory conditions,
LinUCB\,+\,on leads in six scenarios, LinUCB\,+\,off in
software Standard Sprint, and the affinity heuristic in software Maintenance.
The full per-scenario table is reported in \Cref{app:portability}.

\begin{table}[H]
  \centering
  \caption{Main-text summary of cross-scenario portability
    (10 seeds, 8 cycles per scenario). Objective values are lower-is-better
    mean scores for the best human-participatory strategy in each scenario.}
  \label{tab:portability_main}
  \footnotesize
  \setlength{\tabcolsep}{4pt}
  \begin{adjustbox}{max width=0.9\linewidth}
  \begin{tabular}{@{}lllr@{}}
    \toprule
    \textbf{Domain} & \textbf{Scenario} & \textbf{Best human-participatory} & \textbf{Objective} \\
    \midrule
    Software & Standard Sprint    & LinUCB\,+\,off & 195.77 \\
    Software & High Complexity    & LinUCB\,+\,on  & 215.15 \\
    Software & Maintenance        & Affinity       & 143.64 \\
    Software & Deadline Crunch    & LinUCB\,+\,on  & 216.62 \\
    Manufacturing & Standard Production & LinUCB\,+\,on & 150.88 \\
    Manufacturing & Quality Crisis      & LinUCB\,+\,on & 170.51 \\
    Manufacturing & Scheduled Stop      & LinUCB\,+\,on & 123.76 \\
    Manufacturing & New Product Ramp-Up & LinUCB\,+\,on & 164.07 \\
    \bottomrule
  \end{tabular}
  \end{adjustbox}
\end{table}

{\bfseries Interpretation.}
The benchmark consistently shows an AI-only efficiency frontier, but
does not support a universal claim that one governed strategy dominates
across all scenarios.

\subsection{Governance Redistributes Collaboration Modes}
\label{sec:results:policy}

{\bfseries Context.}
The KPI tables report governance \emph{outcomes} but not its \emph{mechanism}.

{\bfseries Findings.}
\textit{Main result.} Governance changes \emph{how} work is allocated more
than \emph{how much} performance is obtained.

{\bfseries Evidence.}
\Cref{tab:mode_main} summarises the mode redistribution under LinUCB;
In software, enabling policies raises \textsc{Supervised} execution from
16.9\% to 54.5\% while cutting \textsc{Autonomous} from 55.2\% to 22.1\%;
the corresponding performance cost is a 3.8\% increase in lead time and an
8.8\% reduction in quality, with negligible fatigue change.
Manufacturing shows the same pattern at a smaller magnitude
(\textsc{Supervised}: 19.9\%\,$\to$\,52.7\%; lead time +0.8\%; quality $-$2.4\%).
The full appendix table additionally reports cost and regret
(\Cref{app:modes}).

\begin{table}[H]
  \centering
  \caption{Main-text summary of governance-induced mode redistribution under
    \textsc{LinUCB} (30 seeds, 8 cycles, audited standard scenarios).}
  \label{tab:mode_main}
  \small
  \setlength{\tabcolsep}{3.5pt}
  \begin{adjustbox}{max width=\linewidth}
  \begin{tabular}{@{}llrrrrrr@{}}
    \toprule
    \textbf{Domain} & \textbf{Policies}
      & \textbf{\shortstack{Lead time\\(h/sprint)}}
      & \textbf{Quality}
      & \textbf{Fatigue}
      & \textbf{Human-Only}
      & \textbf{Supervised}
      & \textbf{Autonomous} \\
    \midrule
    Software & Off & 23.13 & 0.502 & 0.734 & 5.3\% & 16.9\% & 55.2\% \\
    Software & On  & 24.00 & 0.458 & 0.743 & 5.2\% & 54.5\% & 22.1\% \\
    \midrule
    Manufacturing & Off & 24.76 & 0.463 & 0.926 & 8.4\% & 19.9\% & 40.1\% \\
    Manufacturing & On  & 24.96 & 0.452 & 0.889 & 8.6\% & 52.7\% & 14.7\% \\
    \bottomrule
  \end{tabular}
  \end{adjustbox}
\end{table}

{\bfseries Interpretation.}
The mode redistribution data make the governance mechanism visible: policy
enforcement is primarily an authority reallocation, converting
\textsc{Autonomous} slots into \textsc{Supervised} ones rather than shifting
work back to human-only execution.

\subsection{Human State and Well-Being}
\label{sec:results:human}

{\bfseries Context.}
The KPI tables report efficiency metrics but do not directly surface the
implications for human agents: how fatigued are they, and do AI-heavy
strategies erode skill over time?

{\bfseries Findings.}
\textit{Main result.} AI-only minimises fatigue in the short run,
but it also pushes the system toward deskilling because human practice falls
below the minimum threshold. Fixed-Human does the opposite: it avoids
deskilling but saturates fatigue at $f = 1.000$ in both domains. The learned
strategies sit between these extremes.

{\bfseries Evidence.}
\Cref{tab:human_state_main} summarises the human-state reading of the
focal strategies in the two audited standard scenarios. AI-only achieves the
lowest fatigue in both domains (0.124 in software; 0.358 in manufacturing),
but it removes human practice by construction. Fixed-human does the opposite:
it preserves full human exposure but saturates fatigue at 1.000 in both
domains. LinUCB occupies the middle ground: fatigue remains substantially
above AI-only but below the most human-heavy baselines, while mixed human
involvement is preserved. This is exactly the design space that
\textsc{HAAS} is intended to reveal: which configurations preserve human
capability at an operational cost the organisation may still accept.

\begin{table}[H]
  \centering
  \caption{Main-text summary of human-state trade-offs in the audited standard
    scenarios. Values synthesize the focal strategies from
    \Cref{tab:software_results,tab:mfg_results}.}
  \label{tab:human_state_main}
  \footnotesize
  \setlength{\tabcolsep}{2pt}
  \begin{adjustbox}{max width=\linewidth}
  \begin{tabular}{@{}p{0.18\linewidth}rrp{0.15\linewidth}p{0.40\linewidth}@{}}
    \toprule
    \textbf{Strategy}
      & \textbf{\shortstack{Software\\fatigue}}
      & \textbf{\shortstack{Manufacturing\\fatigue}}
      & \textbf{\shortstack{Human\\involvement}}
      & \textbf{Human-state reading} \\
    \midrule
    AI-only baseline           & 0.124 & 0.358 & None & Lowest short-run fatigue, but no human practice exposure. \\
    LinUCB\,+\,off             & 0.734 & 0.926 & Mixed & Preserves non-zero human involvement at a clear fatigue cost. \\
    LinUCB\,+\,on              & 0.743 & 0.889 & Mixed, governed & Similar middle-ground pattern, with governance retaining mixed human participation. \\
    Affinity heuristic baseline & 0.959 & 0.989 & Mixed & Human-state burden is near saturation despite competitive quality in some settings. \\
    Fixed-human baseline       & 1.000 & 1.000 & Full & Maximum practice exposure, but operationally unsustainable fatigue. \\
    \bottomrule
  \end{tabular}
  \end{adjustbox}
\end{table}

{\bfseries Interpretation.}
The human-state results make the benchmark's core trade-off explicit:
short-run fatigue reduction and long-run capability preservation do not point
to the same operating point.

\subsection{Component Ablation}
\label{sec:results:ablation}

{\bfseries Context.}
To separate the contribution of the main architectural layers more explicitly,
we ran a compact LinUCB-centred ablation on the two audited standard
scenarios, using 10 seeds and the same 8-cycle horizon
(\Cref{tab:ablation_main}).
The comparison removes the policy layer, trust dynamics, deskilling
feedback, or fatigue dynamics one at a time while keeping the rest of the
configuration fixed.

{\bfseries Findings.}
\textit{Main result.} The reported trade-off is driven mainly by the policy
layer and fatigue dynamics; deskilling behaves as a longer-horizon signal in
the present benchmark.

{\bfseries Evidence.}
The ablation yields three clear conclusions.
First, removing the policy layer improves regret in both domains
(software: 11.01 $\rightarrow$ 6.94; manufacturing: 11.32 $\rightarrow$ 9.07),
confirming that the \texttt{PolicyEngine} is a substantive source of the
reported trade-off rather than a cosmetic wrapper.
Second, removing fatigue dynamics sharply improves regret and collapses
fatigue, indicating that human-state coupling materially shapes the benchmark
behaviour.
Third, removing deskilling feedback has negligible effect over 8 cycles,
which suggests that deskilling is represented architecturally but remains a
long-horizon signal in the present benchmark rather than a short-run driver
of the headline results.
Trust dynamics have a smaller but still non-zero effect, mainly through
time/fatigue regularisation.
The full ablation table, including lead time and cost, is reported in
\Cref{app:ablation}.

\begin{table}[H]
  \centering
  \caption{Main-text summary of the component ablation
    (\textsc{LinUCB\,+\,on}, 10 seeds, 8 cycles). Lower regret and fatigue
    are better.}
  \label{tab:ablation_main}
  \small
  \setlength{\tabcolsep}{4pt}
  \begin{adjustbox}{max width=\linewidth}
  \begin{tabular}{@{}llrrr@{}}
    \toprule
    \textbf{Domain} & \textbf{Configuration}
      & \textbf{Quality}
      & \textbf{Fatigue}
      & \textbf{Cum.\ regret} \\
    \midrule
    Software & Full model          & 0.456 & 0.833 & 11.01 \\
    Software & No policy layer     & 0.493 & 0.925 &  6.94 \\
    Software & No fatigue dynamics & 0.512 & 0.059 &  7.96 \\
    Software & No trust dynamics   & 0.465 & 0.752 & 10.63 \\
    Software & No deskilling       & 0.456 & 0.833 & 11.01 \\
    \midrule
    Manufacturing & Full model          & 0.451 & 0.886 & 11.32 \\
    Manufacturing & No policy layer     & 0.453 & 0.999 &  9.07 \\
    Manufacturing & No fatigue dynamics & 0.516 & 0.072 &  9.85 \\
    Manufacturing & No trust dynamics   & 0.455 & 0.841 & 11.20 \\
    Manufacturing & No deskilling       & 0.451 & 0.886 & 11.32 \\
    \bottomrule
  \end{tabular}
  \end{adjustbox}
\end{table}

{\bfseries Interpretation.}
The ablation supports the claim that the benchmarked behaviour comes from the
interaction between governance and human-state modelling, not from a generic
bandit wrapper alone. Put differently, the policy layer is not intended to
minimise regret in isolation; it exists to trade some short-run efficiency
for governance compliance and a larger set of diagnostically viable operating
points. That role becomes visible in the screen-pass analysis
(\Cref{tab:contract_appendix}), where the governance-enabled configuration
passes the \emph{responsible} screen more often than the unconstrained
variant across all tested manufacturing reward profiles.

\subsection{Benchmark-Internal Contract Sensitivity in Manufacturing}
\label{sec:results:sensitivity}

{\bfseries Context.}
The strategy ranking above is computed under a single four-outcome reward
profile. How does the set of screen-passing configurations change when a
different reward priority is applied?

{\bfseries Findings.}
\textit{Main result.}
The set of screen-passing configurations is reward-profile-dependent.
Governance (\textsc{LinUCB\,+\,on}) consistently passes the
\emph{responsible} screen more often than the unconstrained variant across
all three profiles tested, whereas the \emph{acceptable} screen is
profile-dependent.

{\bfseries Evidence.}
\Cref{tab:contract_main} summarises the manufacturing sweep across
\emph{four-outcome}, \emph{cost+time}, and \emph{quality+cost+time}
profiles.
Under the four-outcome profile, \textsc{LinUCB\,+\,on} passes the
\emph{acceptable} screen in 53\% of seeds and the \emph{responsible} screen
in 53\% of seeds; \textsc{LinUCB\,+\,off} passes acceptable in 13\% and
responsible in 0\%.
The \emph{cost+time} profile narrows viable configurations further and is the
one case where \textsc{LinUCB\,+\,off} passes the \emph{acceptable} screen
more often than the governance-enabled variant; \emph{quality+cost+time}
shifts the screen-passing balance back toward \textsc{LinUCB\,+\,on}.
The full benchmark-metric table is reported in \Cref{app:contract}.

\begin{table}[H]
  \centering
  \caption{Main-text summary of contract sensitivity in manufacturing
    (30 seeds, 8 cycles). \textit{Acceptable} and \textit{Responsible} denote the
    acceptable and responsible screen pass rates.}
  \label{tab:contract_main}
  \small
  \setlength{\tabcolsep}{4pt}
  \begin{adjustbox}{max width=\linewidth}
  \begin{tabular}{@{}llrrr@{}}
    \toprule
    \textbf{Profile} & \textbf{Strategy}
      & \textbf{Objective}
      & \textbf{\shortstack{Accept-\\able}}
      & \textbf{\shortstack{Respons-\\ible}} \\
    \midrule
    Four-outcome        & LinUCB\,+\,off & 147.1 & 0.13 & 0.00 \\
    Four-outcome        & LinUCB\,+\,on  & 151.3 & 0.53 & 0.53 \\
    Cost\,+\,time       & LinUCB\,+\,off & 133.8 & 1.00 & 0.00 \\
    Cost\,+\,time       & LinUCB\,+\,on  & 152.6 & 0.33 & 0.33 \\
    Quality\,+\,cost\,+\,time & LinUCB\,+\,off & 147.8 & 0.03 & 0.00 \\
    Quality\,+\,cost\,+\,time & LinUCB\,+\,on  & 151.5 & 0.50 & 0.50 \\
    \bottomrule
  \end{tabular}
  \end{adjustbox}
\end{table}

{\bfseries Interpretation.}
The result demonstrates a key property of the \textsc{HAAS} benchmarking
engine: it can expose how the \texttt{PolicyEngine} reshapes the set of
jointly acceptable operating points under any given reward contract,
making the interaction between governance configuration and reward priority
an analysable system-level quantity rather than an implicit design assumption.

%% ================================================================
\section{Discussion}
\label{sec:discussion}
%% ================================================================

This section interprets what the benchmark shows for governed Human--AI work
design, considers how far the cognitive instrument may transfer across
domains, and clarifies both the main limits of the present study and the next
steps needed to extend it.

\subsection{Implications for Work Design}
\label{sec:discussion:design}

The benchmark suggests a practical reading of Human--AI work design.
In many real settings, the question is not whether to maximise automation at
all costs, but how much authority can be delegated without violating
oversight, safety, or capability-retention requirements.

From that perspective, the results support a \emph{graduated autonomy}
approach rather than a single dominant policy. AI-only defines the efficiency
frontier, but governance changes the decision problem by making some degree of
human participation non-negotiable. The resulting cost of governance is best
read as the price of retaining authority and oversight: more supervised work,
slightly slower execution, and in some cases lower quality.

That is the practical role of \textsc{HAAS}: an ex-ante design and
benchmarking artefact rather than an autonomous controller. Organisations can
encode local policy limits, run the benchmark under alternative reward
contracts, and read the resulting collaboration-mode redistribution as the
operational signature of each governance configuration before committing to
one in practice.

The five-dimension cognitive instrument also shows promising cross-domain
stability: applied without modification to both software engineering and
manufacturing tasks, it produced sensible affinity rankings in both domains.
The current evidence is stronger for transferability of the \emph{schema} than
for any specific calibrated weight vector; domain customisation remains
localised to weight calibration (\Cref{eq:affinity}) and subtask catalogue
construction.

\subsection{Governance as a Tunable Design Variable}
\label{sec:discussion:gov_ladder}

The Governance Ladder results refine the paper's central argument in two
important ways.
First, they quantify what the binary comparison only implies: each step up the
ladder produces a predictable authority transfer away from
\textsc{Autonomous} execution and toward \textsc{Supervised} or
human-owned execution, with the magnitude and performance consequence of
that transfer determined by domain structure.
This confirms that the policy layer is not merely symbolic; it exerts a
direct and measurable effect on collaboration patterns.

Second, the results show that the effect of governance is not monotonic in
performance terms.
Greater control does not automatically produce better outcomes, nor does it
uniformly degrade them.
In the software benchmark, stronger governance mainly acts as operational
friction, with the best 8-cycle result remaining at L0 and the upper end of
the ladder becoming clearly over-restrictive.
In manufacturing, by contrast, stronger governance can improve both the main
objective and fatigue, with L3 emerging as the best standard-scenario
operating point --- a \emph{workload-buffering} effect in which supervised
collaboration redistributes physical and cognitive load away from the human
operator, reducing fatigue while maintaining or improving cost efficiency.
The practical implication is that useful governance must be calibrated to the
structure and risk profile of the work rather than applied as a uniform
template.

The portability results further strengthen this reading.
Because L0 and L3 each win four scenarios, the experiment does not support a
single universally optimal governance setting.
What it supports instead is a contextual view: different scenarios absorb
different amounts of authority control before the cost of intervention
outweighs the benefit.
This shifts the design question from ``should governance be enabled?'' to
``how much governance can this operating context productively sustain?''

The long-horizon check adds a further qualification.
At 16 cycles, moderate governance (L2) improves over L0 in both audited
standard scenarios, while very strong governance (L4) remains
over-restrictive.
This suggests that part of the short-run cost of governance reflects learning
friction rather than permanent inefficiency.
Given enough interaction cycles, the learner can adapt within the governed
action space and recover part of that initial penalty.
Governance should therefore be evaluated not only in terms of immediate
throughput loss, but also in terms of how it shapes long-run learning under
operational constraints.

\begin{figure}[H]
  \centering
  \includegraphics[width=0.96\linewidth,height=0.55\textheight,keepaspectratio]{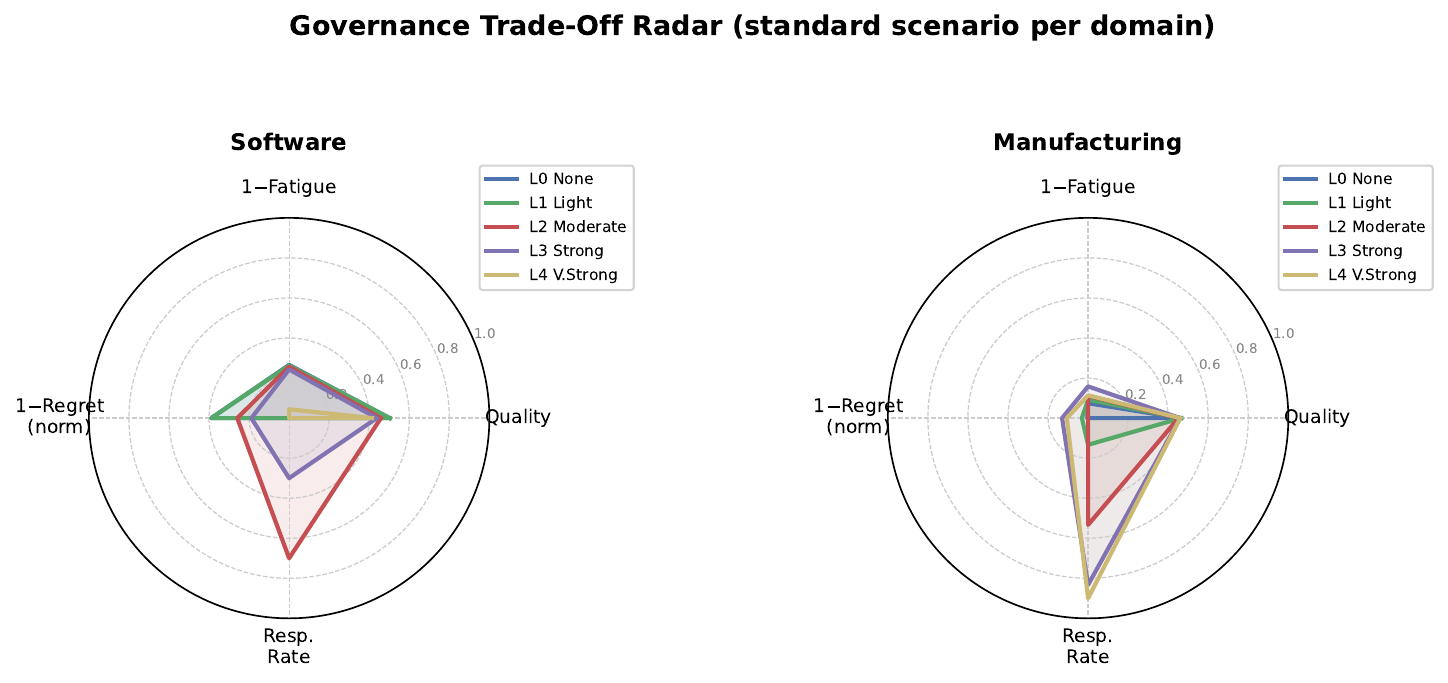}
  \caption{Multi-dimensional trade-off radar for L0--L4
             (standard scenario per domain, 30 seeds).
             Colours: L0 blue, L1 green, L2 red, L3 violet, L4 gold.
             Axes: quality, $1\!-\!\text{fatigue}$,
             $1\!-\!\text{regret}$ (normalised), responsible-strategy rate.
             Larger polygon = better overall profile.
             Manufacturing: L3--L4 uniformly dominate.
             Software: L2 provides the most balanced profile.}
  \label{fig:gov_ladder_radar}
\end{figure}

\Cref{fig:gov_ladder_radar} summarises the multi-dimensional trade-off
in a single view, plotting quality, $1\!-\!\text{fatigue}$, $1\!-\!\text{regret}$
(normalised), and responsible-strategy rate for all five levels (L0--L4) on each domain.
In manufacturing, the L3/L4 profiles expand uniformly relative to L0 across
all four axes; L1 and L2 occupy intermediate positions, confirming the
cumulative structure of the ladder.
In software, L0 leads on quality and regret but has no responsible-strategy
coverage; L1 adds modest governance with minimal quality cost, while L2
achieves the most balanced polygon and remains the pragmatic operating
recommendation.

Taken together, these findings support a bounded claim.
The contribution of the Governance Ladder is not the discovery of a single
best governance level.
Rather, it is the demonstration that governance severity can be
parameterised, benchmarked, and analysed as a controllable dimension of
system behaviour---precisely the kind of evidence needed to study responsible
human-AI allocation as an engineering problem rather than as a binary
policies-on versus policies-off comparison.

\subsection{Limitations}
\label{sec:discussion:limits}

The main limitations are straightforward.
(\emph{i})~\textsc{HAAS} is a simulation framework; human-state parameters
($\beta_f$, $\beta_r$, $\lambda$, trust and deskilling coefficients) are
literature-anchored heuristic defaults, not empirical estimates.
Results should be read as comparative evidence about governance trade-offs
under explicit assumptions, not as deployment forecasts for a specific
organisation.
(\emph{ii})~The benchmark uses a single human-AI dyad over an 8-cycle
horizon chosen to reflect a realistic operational window (two quarters of
weekly sprints, or a standard manufacturing campaign); multi-actor
coordination and deskilling dynamics (which the ablation confirms are
negligible over 8 cycles) remain outside the current scope.
To assess whether this horizon is sufficient for bandit learning to dominate
the heuristic warm-start, we ran an identical sweep at 16 cycles
(30 seeds, both domains).
Strategy rankings were unchanged in 7 of 8 scenarios, and the
lead-time gap between \textsc{LinUCB\,+\,on} and the AI-only
baseline narrowed by 1--2 hours per cycle across all scenarios
(e.g., from $+11.8$\,h to $+9.3$\,h in \emph{Deadline Crunch}),
confirming that the bandit is acquiring subtask-level preferences beyond the
initialisation prior rather than simply replaying the warm-start heuristic.
The 8-cycle results reported in the paper are therefore conservative:
longer horizons favour the learning strategies.
(\emph{iii})~Outcome parameters and cost rates are domain-instantiated rather
than externally calibrated.  The default benchmark uses a parameterised AI
agent for full reproducibility; however, the framework ships
production-ready backends for Groq (Llama-3.1-8b-instant), Google Gemini
(gemini-2.5-flash), Anthropic Claude, and local Ollama models.  Preliminary
experiments with Groq and Gemini across the software and manufacturing
scenarios confirmed that the key comparative findings---strategy rankings,
governance effects on fatigue reduction, and allocation--quality
trade-offs---replicate under real LLM execution; absolute quality levels
showed modest downward shifts ($\Delta q \approx 0.05$--$0.06$) consistent
with model-specific response variation rather than structural disagreement.
Real deployments may still exhibit additional latency, quota constraints,
or safety failure modes not fully captured by the parameterised baseline.
(\emph{iv})~Both the policy catalogue (sparse by design) and the cognitive
instrument (derived deductively) lack field validation; expert elicitation
or confirmatory factor analysis would be needed before deployment in novel
domains.
Taken together, these limits define the paper's scope as a
\emph{simulation-based system and benchmark study}: it supports comparative
reasoning under explicit assumptions, not direct deployment prediction.

\subsection{Future Work}
\label{sec:discussion:future}

The next steps follow directly from those limits.
Priority extensions include:
(\emph{i}) field validation with human participants or structured expert
elicitation to calibrate both the cognitive instrument and the human-state
coefficients;
(\emph{ii}) systematic live-model benchmarks building on the Groq and Gemini
integrations already validated in the platform, extending the preliminary
single-seed alignment results to multi-seed, multi-domain runs and
quantifying latency, cost, and quality deviations relative to the
parameterised baseline across the full strategy catalogue;
(\emph{iii}) dynamic learning of affinity weights and richer reward contracts
from outcome data, reducing manual calibration;
(\emph{iv}) extension to multi-human, multi-AI team settings where assignment
interdependencies create coordination effects; and
(\emph{v}) richer policy libraries, longer horizons, and broader learner
families, including discounted and non-stationary bandits, where adaptation
has more opportunity to dominate heuristics;
(\emph{vi}) adaptive governance as a meta-policy layer that adjusts rule
intensity dynamically in response to operational context, accumulated
fatigue, or detected scenario type; and
(\emph{vii}) a structured domain-adaptation pathway covering expert
elicitation of subtask scores, empirical calibration of human-state
coefficients from task-logging and NASA-TLX data, and formalisation of
site-specific governance constraints to enable deployment beyond simulation.

%% ================================================================
\section{Conclusion}
\label{sec:conclusion}
%% ================================================================

This paper presented \textsc{HAAS}, an implemented intelligent allocation
framework for governance-constrained Human--AI work. \textsc{HAAS} combines a
five-dimension cognitive instrument, a five-mode collaboration spectrum, and
a policy-aware allocation engine in which a rule-based \texttt{PolicyEngine}
constrains learning-based mode selection. The \textsc{HAAS} benchmark platform turns
that architecture into a configurable artefact for benchmarking alternative
allocation policies before deployment.

The benchmark results support a bounded conclusion. In the audited scenarios,
AI-only defines the efficiency frontier, but the Governance Ladder shows that
the useful level of authority control depends on domain and horizon rather
than following a universal optimum. Stronger governance helps in some settings
(most clearly manufacturing) and imposes avoidable friction in others.

The contribution of \textsc{HAAS} is therefore not to argue for a single best
allocation regime, but to provide a reproducible way to inspect how
governance, learning, and human-state dynamics interact before deployment.
That makes the framework useful as an engineering workbench for comparing
feasible Human--AI operating points under explicit organisational
constraints.

%% ================================================================
%% Software availability
%% ================================================================

\section*{Software availability}

The \textsc{Human--AI Symbiosis Studio} benchmark platform is accessible
for interactive evaluation via a public web deployment (URL provided in
the camera-ready version). The benchmark datasets and aggregated simulation
results underlying the reported figures and tables will be deposited in a
public repository with a permanent DOI upon acceptance.

\section*{Declaration of generative AI and AI-assisted technologies in the manuscript preparation process}

During the preparation of this work, the authors used generative AI tools,
including OpenAI Codex and Claude Code, to support software development tasks
related to the application and simulation environment underlying the proposed
framework, as well as manuscript-organisation and language-refinement
tasks during submission preparation. All AI-assisted outputs were reviewed,
verified, and edited by the authors. The authors take full responsibility for
the content of the manuscript, the implemented system, and the reported
results.

%% ---- Bibliography --------------------------------------------
\bibliographystyle{elsarticle-harv}
\bibliography{references_eswa}

%% ================================================================
%% Appendix
%% ================================================================

\FloatBarrier
\appendix

\section{Cross-Scenario Portability — Full Strategy Table}
\label{app:portability}

\Cref{tab:scenario_portability} extends the portability summary in
\Cref{sec:results:portability} to all four focal conditions for each of the
eight benchmark scenarios (10 seeds, 8 cycles).
AI-only is the efficiency winner in all eight scenarios.
Among human-participatory conditions, \textsc{LinUCB\,+\,on} leads in six
scenarios; \textsc{LinUCB\,+\,off} leads in software \emph{Standard Sprint};
and the affinity heuristic leads in software \emph{Maintenance} --- the only
scenario where a non-learning strategy is competitive on the benchmark
objective.

\captionof{table}{Cross-scenario portability sweep (10 seeds, 8 cycles per scenario).
    Objective values (lower is better); \textbf{bold} = best human-participatory
    result per scenario.}
\label{tab:scenario_portability}
\begin{center}
  \footnotesize
  \setlength{\tabcolsep}{3pt}
  \begin{tabular}{@{}llrrrr@{}}
    \toprule
    \textbf{Domain} & \textbf{Scenario}
      & \textbf{AI-only}
      & \textbf{LinUCB\,+\,off}
      & \textbf{LinUCB\,+\,on}
      & \textbf{Affinity} \\
    \midrule
    Software & Standard Sprint    & \phantom{0}38.63 & \textbf{195.77} & 197.63 & 213.77 \\
    Software & High Complexity    & \phantom{0}44.02 & 250.39 & \textbf{215.15} & 242.19 \\
    Software & Maintenance        & \phantom{0}32.32 & 150.34 & 152.21 & \textbf{143.64} \\
    Software & Deadline Crunch    & \phantom{0}38.71 & 254.97 & \textbf{216.62} & 229.56 \\
    \midrule
    Manufacturing & Standard Production & \phantom{0}64.88 & 169.79 & \textbf{150.88} & 164.92 \\
    Manufacturing & Quality Crisis      & \phantom{0}83.09 & 189.87 & \textbf{170.51} & 187.54 \\
    Manufacturing & Scheduled Stop      & \phantom{0}42.22 & 126.75 & \textbf{123.76} & 125.15 \\
    Manufacturing & New Product Ramp-Up & \phantom{0}66.35 & 172.18 & \textbf{164.07} & 171.82 \\
    \bottomrule
    \multicolumn{6}{@{}p{\linewidth}@{}}{\footnotesize
      $n = 10$ seeds per scenario; values are cross-seed mean objective values.} \\
  \end{tabular}
\end{center}

\section{Governance-Induced Mode Redistribution}
\label{app:modes}

\Cref{tab:mode_appendix} quantifies the collaboration-mode shift when
governance is enabled in the two audited standard scenarios
(\textsc{LinUCB}, 30 seeds, 8 cycles).
The dominant effect is conversion of \textsc{Autonomous} assignments into
\textsc{Supervised} ones; \textsc{Human-Only} share does not grow, confirming
that the \texttt{PolicyEngine} redirects AI autonomy into supervised
collaboration rather than replacing AI with human labour.

\captionof{table}{Collaboration-mode redistribution under \textsc{LinUCB} with
    governance off vs.\ on (30 seeds, 8 cycles, audited standard scenarios).}
\label{tab:mode_appendix}
\begin{center}
  \footnotesize
  \setlength{\tabcolsep}{2pt}
  \begin{tabular}{@{}llrrrrrrrr@{}}
    \toprule
    \textbf{Domain} & \textbf{Policies}
      & \textbf{\shortstack{Lead time\\(h/sprint)}}
      & \textbf{Quality}
      & \textbf{Cost (€)}
      & \textbf{Fatigue}
      & \textbf{\shortstack{Cum.\\regret}}
      & \textbf{\shortstack{Human-\\Only}}
      & \textbf{Supervised}
      & \textbf{\shortstack{Auto-\\nomous}} \\
    \midrule
    Software & Off & 23.13 & 0.502 & 645.91 & 0.734 & 10.69 & 5.3\% & 16.9\% & 55.2\% \\
    Software & On  & 24.00 & 0.458 & 739.44 & 0.743 & 12.96 & 5.2\% & 54.5\% & 22.1\% \\
    \midrule
    Manufacturing & Off & 24.76 & 0.463 & 588.54 & 0.926 & 10.94 & 8.4\% & 19.9\% & 40.1\% \\
    Manufacturing & On  & 24.96 & 0.452 & 605.24 & 0.889 & 10.98 & 8.6\% & 52.7\% & 14.7\% \\
    \bottomrule
    \multicolumn{10}{@{}p{\linewidth}@{}}{\footnotesize
      Remaining share distributed across \textsc{Copilot} and \textsc{Peer}.
      Governance on = L2 (Moderate) profile.} \\
  \end{tabular}
\end{center}

\section{Component Ablation}
\label{app:ablation}

\Cref{tab:ablation_appendix} isolates the contribution of each architectural
component by removing it while holding the rest fixed
(10 seeds, standard scenario per domain, 8 cycles, governance on).

\captionof{table}{Component ablation: \textsc{LinUCB\,+\,on}, standard scenario
    per domain (10 seeds, 8 cycles). Lower regret and fatigue are better.}
\label{tab:ablation_appendix}
\begin{center}
  \footnotesize
  \setlength{\tabcolsep}{4pt}
  \begin{tabular}{@{}llrrrrr@{}}
    \toprule
    \textbf{Domain} & \textbf{Configuration}
      & \textbf{\shortstack{Lead time\\(h/sprint)}}
      & \textbf{Quality}
      & \textbf{Fatigue}
      & \textbf{Cost (€)}
      & \textbf{Cum.\ regret} \\
    \midrule
    Software & Full model            & 24.67 & 0.456 & 0.833 & 790.5 & 11.01 \\
    Software & No policy layer       & 25.07 & 0.493 & 0.925 & 783.1 &  6.94 \\
    Software & No fatigue dynamics   & 24.74 & 0.512 & 0.059 & 825.0 &  7.96 \\
    Software & No trust dynamics     & 22.46 & 0.465 & 0.752 & 679.4 & 10.63 \\
    Software & No deskilling         & 24.67 & 0.456 & 0.833 & 790.4 & 11.01 \\
    \midrule
    Manufacturing & Full model            & 24.94 & 0.451 & 0.886 & 603.5 & 11.32 \\
    Manufacturing & No policy layer       & 27.00 & 0.453 & 0.999 & 679.2 &  9.07 \\
    Manufacturing & No fatigue dynamics   & 24.74 & 0.516 & 0.072 & 617.5 &  9.85 \\
    Manufacturing & No trust dynamics     & 22.96 & 0.455 & 0.841 & 541.1 & 11.20 \\
    Manufacturing & No deskilling         & 24.94 & 0.451 & 0.886 & 603.5 & 11.32 \\
    \bottomrule
    \multicolumn{7}{@{}p{\linewidth}@{}}{\footnotesize
      ``No policy layer'' = \textsc{LinUCB\,+\,off}.
      Near-zero fatigue in ``No fatigue dynamics'' confirms that fatigue
      coupling is the primary driver of the reported well-being trade-off.} \\
  \end{tabular}
\end{center}

\section{Contract Sensitivity — Manufacturing Domain}
\label{app:contract}

\Cref{tab:contract_appendix} shows how the set of screen-passing
configurations changes across three reward profiles in manufacturing
(30 seeds, standard scenario, 8 cycles).
The governance-enabled variant (\textsc{LinUCB\,+\,on}) consistently passes
the \emph{responsible} diagnostic screen more often than the unconstrained
variant across all three tested profiles. For the \emph{acceptable} screen,
the advantage depends on the reward contract: governance improves pass rates
under the four-outcome and quality+cost+time profiles, whereas
\textsc{LinUCB\,+\,off} performs better under cost+time. This demonstrates
that the \texttt{PolicyEngine} can expand the jointly viable operating
region, but that the effect depends on which diagnostic screen and reward
contract are being applied.

\captionof{table}{Manufacturing contract sensitivity: mean benchmark metrics and
    governance screen rates by reward profile (30 seeds, 8 cycles).
    \textit{Acceptable} = fraction of seeds passing the \emph{acceptable} screen;
    \textit{Responsible} = fraction passing the \emph{responsible} screen.}
\label{tab:contract_appendix}
\begin{center}
  \footnotesize
  \setlength{\tabcolsep}{2pt}
  \begin{adjustbox}{max width=\linewidth}
  \begin{tabular}{@{}lllrrrrrrr@{}}
    \toprule
    \textbf{Profile} & \textbf{Strategy}
      & \textbf{Objective}
      & \textbf{Quality}
      & \textbf{Fatigue}
      & \textbf{Cost (€)}
      & \textbf{\shortstack{Lead time\\(h/sprint)}}
      & \textbf{\shortstack{Cum.\\regret}}
      & \textbf{\shortstack{Accept-\\able}}
      & \textbf{\shortstack{Respons-\\ible}} \\
    \midrule
    \multirow{4}{*}{\shortstack[l]{Four-\\outcome}}
      & AI-only        &  64.9 & 0.496 & 0.358 & 259.5 & 17.4 &  3.5 & 0.00 & 0.00 \\
      & LinUCB\,+\,off & 147.1 & 0.463 & 0.926 & 588.5 & 24.8 & 10.9 & 0.13 & 0.00 \\
      & LinUCB\,+\,on  & 151.3 & 0.452 & 0.889 & 605.2 & 25.0 & 11.0 & 0.53 & 0.53 \\
      & Affinity       & 164.9 & 0.493 & 0.989 & 659.7 & 26.9 & 10.2 & 0.00 & 0.00 \\
    \midrule
    \multirow{4}{*}{\shortstack[l]{Cost\,+\\time}}
      & AI-only        &  64.9 & 0.496 & 0.358 & 259.5 & 17.4 & — & 0.00 & 0.00 \\
      & LinUCB\,+\,off & 133.8 & 0.464 & 0.850 & 535.1 & 23.5 & — & 1.00 & 0.00 \\
      & LinUCB\,+\,on  & 152.6 & 0.448 & 0.899 & 610.5 & 25.0 & — & 0.33 & 0.33 \\
      & Affinity       & 164.9 & 0.493 & 0.989 & 659.7 & 26.9 & — & 0.00 & 0.00 \\
    \midrule
    \multirow{4}{*}{\shortstack[l]{Quality\,+\\cost\,+time}}
      & AI-only        &  64.9 & 0.496 & 0.358 & 259.5 & 17.4 & — & 0.00 & 0.00 \\
      & LinUCB\,+\,off & 147.8 & 0.462 & 0.930 & 591.3 & 24.8 & — & 0.03 & 0.00 \\
      & LinUCB\,+\,on  & 151.5 & 0.451 & 0.891 & 605.9 & 25.0 & — & 0.50 & 0.50 \\
      & Affinity       & 164.9 & 0.493 & 0.989 & 659.7 & 26.9 & — & 0.00 & 0.00 \\
    \bottomrule
  \end{tabular}
  \end{adjustbox}
  \par\smallskip
  \begin{minipage}{\linewidth}
    \footnotesize\noindent
    Screens are system diagnostics: \textit{Acceptable} requires quality,
    fatigue, deskilling, human-participation, and governance-compliance
    conditions to hold simultaneously.
    \textit{Responsible} additionally requires low monotony, a meaningful share
    of shared modes, limits on fully autonomous high-value work, and minimum
    human retention in high-value and risky tasks.
  \end{minipage}
\end{center}

\section{Seed-Level Dispersion}
\label{app:dispersion}

\captionof{table}{Seed-level dispersion for the pre-specified AI-only vs.\
    \textsc{LinUCB\,+\,off} comparison (median {[}Q1, Q3{]} across 30 seeds).}
\label{tab:dispersion_appendix}
\begin{center}
  \footnotesize
  \begin{tabularx}{\linewidth}{@{}ll>{\raggedright\arraybackslash}X>{\raggedright\arraybackslash}X>{\raggedright\arraybackslash}X>{\raggedright\arraybackslash}X@{}}
    \toprule
    \textbf{Domain} & \textbf{Strategy} & \textbf{\shortstack{Lead time\\(h/sprint)}} & \textbf{Quality} & \textbf{Fatigue} & \textbf{\shortstack{Cum.\\regret}} \\
    \midrule
    Software & AI-only & 16.63 [16.62, 16.63] & 0.505 [0.503, 0.506] & 0.124 [0.124, 0.124] & 4.099 [4.097, 4.103] \\
    Software & LinUCB\,+\,off & 23.12 [23.10, 23.17] & 0.501 [0.496, 0.508] & 0.734 [0.730, 0.737] & 10.671 [10.562, 10.781] \\
    Manufacturing & AI-only & 17.38 [17.38, 17.39] & 0.497 [0.494, 0.499] & 0.358 [0.358, 0.358] & 3.482 [3.452, 3.506] \\
    Manufacturing & LinUCB\,+\,off & 24.72 [24.58, 24.94] & 0.462 [0.457, 0.467] & 0.927 [0.905, 0.941] & 10.912 [10.700, 11.094] \\
    \bottomrule
  \end{tabularx}
\end{center}

\section{Simulator Calibration Workflow}
\label{app:calibration}

\captionof{table}{Calibration workflow implemented in the repository tuning runner.}
\label{tab:tuning_appendix}
\begin{center}
  \footnotesize
  \begin{tabularx}{0.92\linewidth}{@{}>{\raggedright\arraybackslash}p{4.0cm}>{\raggedright\arraybackslash}X@{}}
    \toprule
    \textbf{Element} & \textbf{Configuration} \\
    \midrule
    Candidate family & Affinity weights, four-outcome reward weights, well-being thresholds, exploration setting, discount factor $\gamma$, and LinUCB confidence parameter $\alpha$. \\
    Search space & Affinity jitter 0.28; reward and well-being jitter 0.35; threshold jitter 0.20; $\gamma \in [0.90, 0.97]$; $\alpha \in [0.35, 1.60]$. \\
    Ranking rule & Feasible-first: constraint satisfaction prioritised before objective improvement. \\
    Purpose & Internal simulator calibration for coherent benchmark defaults, not field validation. \\
    \bottomrule
  \end{tabularx}
\end{center}

\clearpage
\section{Simulation Parameter Reference}
\label{app:params}

\captionof{table}{Key simulation parameters and default values.}
\label{tab:params}
\begin{center}
  \footnotesize
  \begin{tabularx}{0.92\linewidth}{@{}>{\raggedright\arraybackslash}p{3.2cm}>{\raggedright\arraybackslash}Xr@{}}
    \toprule
    \textbf{Parameter} & \textbf{Description} & \textbf{Default} \\
    \midrule
    $C$ (UCB1)                  & Exploration constant            & 1.5  \\
    $\gamma$ (D-UCB)            & Discount factor                 & 0.97 \\
    $\alpha$ (LinUCB)           & Confidence width                & 0.8  \\
    $T_{\mathrm{explore}}$      & Heuristic warm-up sprints       & 3    \\
    $\delta_{\mathrm{peer}}$    & Peer mode balance threshold     & 0.20 \\
    $f_{\mathrm{hybrid}}$       & Hybrid trigger fatigue          & 0.35 \\
    $\sigma_{\max}$             & Copilot AI share upper bound    & 0.55 \\
    $\beta_f$                   & Base fatigue rate (h$^{-1}$)   & 0.07 \\
    $\beta_r$                   & Fatigue recovery rate           & 0.12 \\
    $\lambda$                   & Chronic fatigue fraction        & 0.18 \\
    $\rho_{\mathrm{desk}}$      & Deskilling AI ratio threshold   & 0.80 \\
    $(w_r, w_\tau, w_c, w_a, w_h)$ & Five-dimension affinity weights & (0.35, 0.25, 0.20, 0.10, 0.10) \\
    $(w_q, w_t, w_{\mathrm{co}}, w_w)$ & Main four-outcome reward weights & (0.30, 0.20, 0.10, 0.40) \\
    \bottomrule
  \end{tabularx}
\end{center}

\end{document}